\documentclass[sigconf,screen,nonacm]{acmart}
\AtBeginDocument{%
  }

\setcopyright{cc}
\setcctype{by}
\usepackage{graphicx}
\usepackage{subcaption}
\usepackage{pdfpages}
\usepackage[english]{babel}
\usepackage[autostyle, english = american]{csquotes}
\usepackage{multirow}
\usepackage{tabularx}
\usepackage{booktabs}
\usepackage{graphicx}
\usepackage{array}
\usepackage{arydshln}
\begin{document}





\title[Designing Social Robots for Inclusive Child Wellbeing Assessment]{Designing Social Robots for Inclusive Child Wellbeing Assessment: Insights from Communities Supporting Developmental Language Disorder and Forced Migration}


\author{Fethiye Irmak Dogan}
\authornote{Equal first authorship.}
\orcid{0000-0002-1733-7019}
\affiliation{%
  \institution{Department of Computer Science \& Technology, University of Cambridge}
  \city{Cambridge}
  \country{UK}}
\email{fid21@cam.ac.uk}

\author{Yue Lou}
\authornotemark[1]
\orcid{0009-0006-1494-782X}
\affiliation{%
  \institution{Department of Information Technology, Uppsala University}
  \city{Uppsala}
  \country{Sweden}}
\email{yue.lou@it.uu.se}

\author{Alva Markelius}
\orcid{0009-0003-4580-9997}
\affiliation{%
  \institution{Department of Computer Science \& Technology, University of Cambridge}
  \city{Cambridge}
  \country{UK}}
\email{ajkm4@cam.ac.uk}

\author{Emma Geijer-Simpson}
\orcid{0009-0001-6132-8638}
\affiliation{%
  \institution{Department of Public Health and Caring Sciences, Uppsala University}
  \city{Uppsala}
  \country{Sweden}}
\email{emma.geijersimpson@uu.se}

\author{Gustaf Gredebäck}
\orcid{0000-0003-3046-0043}
\affiliation{%
  \institution{Department of Psychology, Uppsala University}
  \city{Uppsala}
  \country{Sweden}}
\email{gustaf.gredeback@psyk.uu.se}

\author{Tamsin Jane Ford}
\orcid{0000-0001-5295-4904}
\affiliation{%
  \institution{Department of Psychiatry, University of Cambridge}
  \city{Cambridge}
  \country{UK}}
\email{tjf52@medschl.cam.ac.uk}

\author{Ginevra Castellano}
\orcid{0000-0002-2841-6791}
\authornote{Equal senior authorship.}
\affiliation{%
  \institution{Department of Information Technology, Uppsala University}
  \city{Uppsala}
  \country{Sweden}}
\email{ginevra.castellano@it.uu.se}

\author{Hatice Gunes}
\orcid{0000-0003-2407-3012}
\authornotemark[2]
\affiliation{%
  \institution{Department of Computer Science \& Technology, University of Cambridge}
  \city{Cambridge}
  \country{UK}}
\email{hg410@cam.ac.uk}

\author{Georgina Warner}
\orcid{0000-0002-7850-9136}
\authornotemark[2]
\affiliation{%
  \institution{Department of Public Health and Caring Sciences, Uppsala University}
  \city{Uppsala}
  \country{Sweden}}
\email{georgina.warner@uu.se}

\author{Jenny L. Gibson}
\orcid{0000-0002-6172-6265}
\authornotemark[2]
\affiliation{%
  \institution{Faculty of Education, University of Cambridge}
  \city{Cambridge}
  \country{UK}}
\email{jlg53@cam.ac.uk}

\renewcommand{\shortauthors}{Dogan et al.}

\begin{abstract}
Assessing children's wellbeing and mental health can be particularly challenging for children experiencing communication barriers, such as children with Developmental Language Disorder (DLD) and children with forced migration backgrounds. During the assessment process, traditional self-report questionnaires place substantial demands on language comprehension and verbal expression. In this context, social robots have emerged as a promising tool for supporting wellbeing assessment without solely relying on self-report questionnaires, yet limited research has examined how such interactions can be designed to be inclusive, appropriate, and ethically acceptable for children with diverse communication needs. To address this gap, we created candidate child--robot interaction activities as design probes and conducted focus groups with parents and professionals supporting children with DLD and children with forced migration backgrounds. Through thematic analysis, we identified considerations relating to robot role and capabilities, interactional dynamics, individual differences, and child agency, alongside population-specific considerations shaped by children's communication needs and lived experiences. Based on these findings, we derive a set of ethical and inclusive design recommendations for robot-mediated wellbeing assessment. By foregrounding these considerations and recommendations, this work contributes design guidance for inclusive robot-mediated wellbeing assessments for children with diverse communication needs.

\end{abstract}




\maketitle

\section{Introduction}

Child wellbeing and mental health assessment targets identifying children's strengths, needs, and early signs of difficulty, thereby informing timely educational, clinical, and psychosocial support~\cite{microwellbeing, deighton2014measuring, connors2022advancing}. Yet, because children's wellbeing and mental health are multidimensional concepts, their assessment is inherently difficult and complex~\cite{ben2014multifaceted}. 
This complexity may be further compounded for children who experience barriers to communication, such as children with Developmental Language Disorder (DLD) and children with forced migration backgrounds. DLD, affecting approximately one in fourteen children at school entry, is characterised by difficulties in understanding and/or using spoken language~\cite{nidcd_dld_2023}. Similarly, children who have experienced forced migration may face linguistic barriers that can affect their ability to communicate wellbeing needs and access appropriate support~\cite{fazel2018preventive}. Although these populations are highly heterogeneous, both may experience difficulties communicating their experiences, emotions, and wellbeing needs through traditional assessment approaches.

Given the relationship between communicative participation and wellbeing outcomes~\cite{duinmeijer2025language}, assessing wellbeing in children who experience communication barriers can benefit from multimodal approaches. In the assessment process, self-report questionnaires, while commonly used, can be particularly limited: language impairments can affect how children with DLD understand and respond to questions, while disrupted schooling and learning a new language may create additional barriers for children with forced migration backgrounds~\cite{microwellbeing}. 
In this context, social robots have emerged as a promising and accessible tool in recent years~\cite{10.1145/3722123,abbasi2024analysing,9900843}. As embodied and interactive agents, they can support multimodal interactions that combine speech, non-verbal behaviour, facial expressions, gestures, and social engagement~\cite{FONG2003143, 10.3389/fnbot.2023.1084000}. Building on these capabilities, recent research has explored the use of social robots to encourage discussions about emotions and personal experiences~\cite{9900843}, administer wellbeing questionnaires~\cite{10.1145/3722123}, and elicit verbal and non-verbal behavioural indicators relevant to children's wellbeing~\cite{abbasi2024analysing}. Together, these developments highlight the potential of social robots as a tool for supporting wellbeing assessment and motivate further investigation into how such interactions can be designed for diverse groups of children.

Despite this growing interest, existing work has primarily focused on establishing the feasibility of robot-mediated wellbeing assessment and demonstrating the ability of social robots to elicit wellbeing-related information~\cite{10.1145/3722123,abbasi2024analysing,9900843}. Comparatively less attention has been paid to how such interactions should be designed for children with diverse communication, linguistic, and cultural needs. Consequently, \textbf{there remains a limited understanding of how robot-mediated wellbeing assessment can be implemented in ways that are appropriate, inclusive, and ethically acceptable for children with DLD and children with forced migration backgrounds.} 
Addressing this gap is important for ensuring that these assessments are accessible and responsive to the needs of diverse child populations.

In this paper, we explore design considerations for robot-mediated wellbeing assessment for children aged 8--11 years through the perspectives of communities supporting children with DLD and forced migration backgrounds. To gather these perspectives, we conducted focus groups with parents and professionals from these communities. We first developed a set of candidate child--robot interaction activities targeting language use, social interaction, non-verbal communication, and emotional understanding and expression. These activities then served as design probes through which participants could reflect on the appropriateness, inclusiveness, and ethical implications of robot-mediated wellbeing assessment. 
We selected DLD and forced migration as two illustrative contexts of diverse communication needs because they reflect different pathways through which communication may affect wellbeing assessment: DLD involves neurodevelopmental differences in understanding and/or using spoken language, whereas forced migration may involve linguistic, cultural, and psychosocial barriers. These contexts were also aligned with the expertise of the interdisciplinary project team and enabled us to explore how such considerations may manifest in groups whose communication-related needs arise for different reasons and may be experienced in different ways.
In doing so, \textbf{this work contributes to the broader goal of informing the design of social robots for wellbeing assessment that are inclusive of diverse communication needs.} Specifically, we ask the following research questions:

\begin{itemize}
    \item \textbf{RQ1.} What considerations do stakeholders (e.g., parents and professionals) identify for robot-mediated wellbeing assessment, particularly for children with diverse communication needs?
    \item \textbf{RQ2.} How do these considerations manifest in the contexts of children with DLD and children with forced migration backgrounds?
    \item \textbf{RQ3.} How can these considerations be translated into inclusive and ethical design recommendations for social robots intended to support children's wellbeing assessment?
\end{itemize}


To address these research questions, we analysed the focus group discussions using thematic analysis. Our findings revealed considerations relating to establishing the role and capabilities of the robot, recognising the importance of interactional dynamics, accommodating individual differences, and cultivating child agency during robot-mediated wellbeing assessment. While these considerations were broadly shared across both populations, participants also highlighted needs shaped by children's distinct communication needs and lived experiences. For children with DLD, stakeholders emphasised communication support, adaptation, and adult facilitation, whereas discussions of forced migration underlined cultural and linguistic sensitivity, trust-building, and the need to account for experiences of displacement and belonging. Based on these findings, we present a set of recommendations for robot-mediated wellbeing assessment, emphasising human oversight, community involvement, bias awareness, content safeguards, and careful consideration of robot embodiment and interaction design. Together, these recommendations aim to inform the design of ethical and inclusive robot-mediated wellbeing assessments for children with diverse communication needs.


\section{Related Work}

\subsection{Child Wellbeing Assessment and Communication Barriers}

Children's wellbeing is commonly measured through standardised self-report questionnaires due to their practical efficiency and scalability. These range from tools focused on specific constructs such as mood, anxiety, and depression (e.g., SMFQ \cite{SMFQ}, RCADS \cite{RCADS}), to broader holistic measures capturing multiple wellbeing domains (e.g., KIDSCREEN \cite{ravens2005kidscreen}, KINDL \cite{ravens1998assessing}). While proxy versions completed by parents or teachers are available, children's own reports are considered to have greater face validity for capturing subjective wellbeing and perceived needs \cite{RILEY2004371}. Although children aged 8 and above can generally provide sufficiently valid self-report data, response validity depends on cognitive prerequisites including comprehension, sustained attention, and recall ability \cite{RILEY2004371, Conijn_Smits_Hartman_2020}, as well as susceptibility to response biases, with emotional and abstract items proving particularly challenging \cite{Conijn_Smits_Hartman_2020}.

These limitations are especially relevant for children who experience communication difficulties. Children with DLD already face elevated risk of mental health difficulties compared to their typically developing peers \cite{dld_qol}, yet the very language impairments that contribute to this risk can also interfere with their ability to understand and respond accurately to the questionnaire items designed to assess it \cite{dld_book}. Children from forced migration backgrounds face a distinct but overlapping set of communication barriers. This population already faces heightened mental health challenges compared to their peers, shaped by multiple layers of difficulties across the migration journey, including pre-migration trauma, adverse experiences during their flight such as exploitation and abuse, and post-migration stressors such as acculturative stress and discrimination in the host country \cite{FAZEL2012266, Scharpf_Kaltenbach_Nickerson_Hecker_2021}. Language barriers, unfamiliarity with social norms in the host country, and disruptions to prior schooling \cite{participatory_refugee_2026, Yang_Hu_Dautenhahn_2025, unicef2017education} can together limit children's ability to engage with the demands of standard assessment formats. 

Together, these contexts point to the need for inclusive assessment methods that can accommodate diverse communication needs without reliance on self-report questionnaires. Social robots have emerged as a promising alternative, offering multimodal and interactive engagement to support direct wellbeing assessment, as discussed in the following section.

\subsection{Social Robots for Child Wellbeing Assessment} 

Empirical work on robot-mediated wellbeing assessment with children has grown in recent years, motivated by the limitations of traditional self-report methods and the unique affordances of social robots as interactive assessment partners. Research has explored robots' capacity to elicit children's emotional experiences and personal disclosures, administer standardised wellbeing questionnaires such as the SMFQ \cite{SMFQ} and RCADS \cite{RCADS}, and present pictorial stimuli to elicit verbal narratives as a basis for wellbeing measurement \cite{9900843, 10.1145/3722123}. Longitudinal work has further demonstrated that repeated robot-mediated interactions can support sustained wellbeing monitoring over time \cite{10.1145/3722123}. Beyond verbal responses, social robots support multimodal interaction through voice, gesture, and facial expression, enabling the capture of non-verbal behavioural indicators through both passive observation and structured interaction \cite{affectiveWellbeing, Rasouli_Gupta_Nilsen_Dautenhahn_2022}. Recent work has also begun integrating vision-language models into robot-led assessment, showing moderate feasibility while highlighting persistent challenges around accuracy and demographic bias \cite{11217833}.

A key affordance of robots in this context is their capacity to support self-disclosure. Children have been shown to share sensitive personal information with humanoid robots \cite{secret_sharing, bullying}, a finding that presents both an opportunity for eliciting meaningful wellbeing-related data and a set of ethical considerations that warrant careful attention, particularly when working with vulnerable populations~\cite{safeguarding_2026}.

Despite this growing body of work, existing studies have predominantly focused on typically developing children in controlled research settings. Knowledge of how robot-mediated assessment should be designed to be inclusive and ethically appropriate for children with diverse communication needs remains limited, with stakeholder perspectives on this question similarly underrepresented in the literature. This study addresses this gap by examining inclusive and ethical design for robot-mediated wellbeing assessment with two vulnerable populations: children with DLD and children from forced migration backgrounds.

\subsection{Participatory and Community-Informed Approaches to Wellbeing Assessment and HRI}

Participatory and co-design approaches are now established in the design of wellbeing-related technologies, reflecting a shift from designing \textit{for} children and young people toward designing \textit{with} them, who are treated as experts in their own experience \cite{Miao_Yin_Zhang_Siu_2025}. Within HRI, participatory design has been used to shape wellbeing robots, for example to inform the form, behaviours, and practices of a robotic mental wellbeing coach \cite{Axelsson_Bodala_Gunes_2021}, for robotic roles to address challenges of perinatal depression screening \cite{zhong2026designing, tanqueray2022gender}, repair strategies for robot mistakes made in wellbeing related coaching \cite{axelsson2024oh}, and investigating ethical and socio-technical considerations of wellbeing robotics \cite{axelsson2025owns}. This work establishes participation as a route to wellbeing technologies that fit users' needs, though it has centred on designing interventions rather than on the assessment task. In child-robot interaction more specifically, the embodied presence of robots in children's physical and social environments creates opportunities to support wellbeing while also raising concerns tied to children's fundamental rights, including inclusion, non-discrimination, agency, and participation \cite{Charisi_Sabanovic_Cangelosi_Gomez_2021}. Addressing these concerns has motivated calls to involve children across all stages of design and to extend participation to under-represented communities whose perspectives are often absent from the field \cite{Charisi_Sabanovic_Cangelosi_Gomez_2021}, with some work beginning to engage children as co-researchers in participatory design to explore future social robots for wellbeing \cite{childWellbeingParticipatory}.

A growing body of HRI research engages populations with disabilities and communication differences through participatory and community-based methods. Co-design with disabled higher education students, their representatives, and disability practitioners has been used to envision social robots for advocacy and support, deliberately reframing the robot away from a corrective intervention and toward a tool that amplifies users' agency \cite{Markelius_Bailey_Gibson_Gunes_2025}. Work with adults with intellectual disability shows how participation can move beyond consultation toward sustained partnership, with a participant progressing from research subject to co-researcher and contributing design insight drawn from lived experience \cite{Haidenhofer_Sitbon_Beaumont_Hoogstrate_Korte_2024}. Participatory and community-informed approaches have similarly been applied for children with migration and refugee backgrounds. Teachers have been involved in co-creating robot-mediated activities to promote the inclusion of children with migration backgrounds, first mapping classroom challenges and integration strategies before identifying suitable robot roles \cite{Tozadore_Kuoppamaki_Guneysu_2023}. Professionals working with government-assisted refugee families have likewise shaped the design of robot-assisted language learning, valuing the robot as a non-judgemental partner and a neutral medium through which children can engage with unfamiliar social norms \cite{Yang_Hu_Dautenhahn_2025}. 

Across these studies, participatory and community-informed work in HRI has concentrated on designing interventions, such as coaching, advocacy, integration, and language learning, rather than on designing wellbeing \textit{assessment}. This reflects participatory wellbeing research more broadly, where assessment is seldom the target of participation; in one review, few studies used participatory methods to evaluate wellbeing or intervention impact, in one case by co-developing indicators of wellbeing and belonging with migrant and refugee children \cite{Miao_Yin_Zhang_Siu_2025}. How to design an assessment process that elicits valid wellbeing-related information, and how to do so inclusively for children with diverse communication needs, has therefore received little attention. We address this gap by applying participatory, community-informed methods not to the design of an intervention but to the design of an inclusive, robot-mediated approach to wellbeing assessment, drawing on the perspectives of those who support children with DLD and forced migration backgrounds.


\section{Methodology}

Our methodology forms an initial qualitative, participatory-design phase within the broader MICRO project, which investigates social robots as a route toward more accessible and child-centred wellbeing assessment~\cite{microwellbeing}. The aim of this phase was to elicit stakeholder input to inform the design of future child--robot interaction scenarios for children aged 8 years to 11 years and 11 months, hereafter referred to as aged 8--11 years.  We therefore adopted a participatory, community-informed approach. We first developed a set of candidate child--robot interaction activities targeting verbal, non-verbal, social, and emotional behaviours relevant to wellbeing assessment (Section~\ref{sec:activity_development}). These activities were then used as design probes in focus groups with parents and professionals supporting children with DLD and children with forced migration backgrounds -- see Section~\ref{sec:focus_groups}. The focus group protocol included discussions addressing stakeholders' general expectations, perceived opportunities, concerns, and ethical considerations regarding robot-mediated wellbeing assessment (RQ1), as well as considerations specific to DLD and forced migration contexts (RQ2). The resulting discussions were analysed using thematic analysis to formulate inclusive and ethical design recommendations for future robot-mediated wellbeing assessment (RQ3) -- see Section~\ref{sec:data_analysis}.


\subsection{Activity Development}
\label{sec:activity_development}
\subsubsection{Interdisciplinary Design Process}

The initial activity set was developed through an interdisciplinary collaboration involving researchers specialising in robotics, human–robot interaction (HRI), affective computing, neurodiversity and developmental psychology, and child and adolescent psychiatry. The design process aimed to identify child–robot interaction activities that could support the assessment of children's wellbeing while remaining engaging, developmentally appropriate, and accessible to diverse populations. The HRI and robotics expertise informed the technical feasibility and interaction design of the proposed activities, relying on prior experience and expertise in utilising social robots for wellbeing assessment~\cite{9900843, 10.1145/3722123}, while developmental psychology and child psychiatry perspectives ensured that activities aligned with children's cognitive, social, emotional, and communicative abilities. Expertise in affective computing contributed to the design of activities capable of eliciting and capturing verbal and non-verbal indicators of emotional state and wellbeing.

Through a series of interdisciplinary discussions and iterations, informed by prior literature on child wellbeing~\cite{pollard2003child}, socio-emotional development and communication difficulties~\cite{levickis2018language}, and child--robot interaction~\cite{polycarpou2016don, 10.1007/978-3-031-24670-8_3, 9900843, 11217833}, the team identified three domains, namely, (i) language and communication, (ii) social and non-verbal interaction, and (iii) emotional understanding and expression as key domains of focus, as they represent important behavioural and socio-emotional dimensions of children's wellbeing~\cite{10.1007/978-3-031-24670-8_3}. Particular attention was paid to ensuring that the resulting approach would be appropriate for children aged 8--11 years, including children with Developmental Language Disorder (DLD) and children with forced migration backgrounds.


\begin{figure*}[t!]
    \centering

    \begin{subfigure}{0.24\textwidth}
        \includegraphics[width=\linewidth]{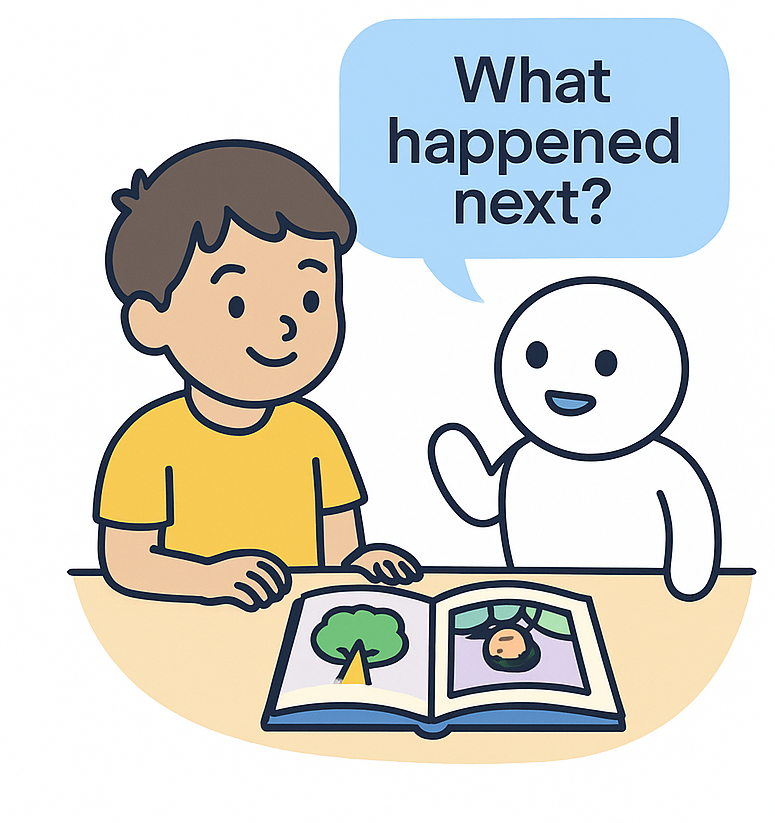}
        \caption{Structured Storytelling}
        \label{activity_story}
    \end{subfigure}
    \hfill
    \begin{subfigure}{0.24\textwidth}
        \includegraphics[width=\linewidth]{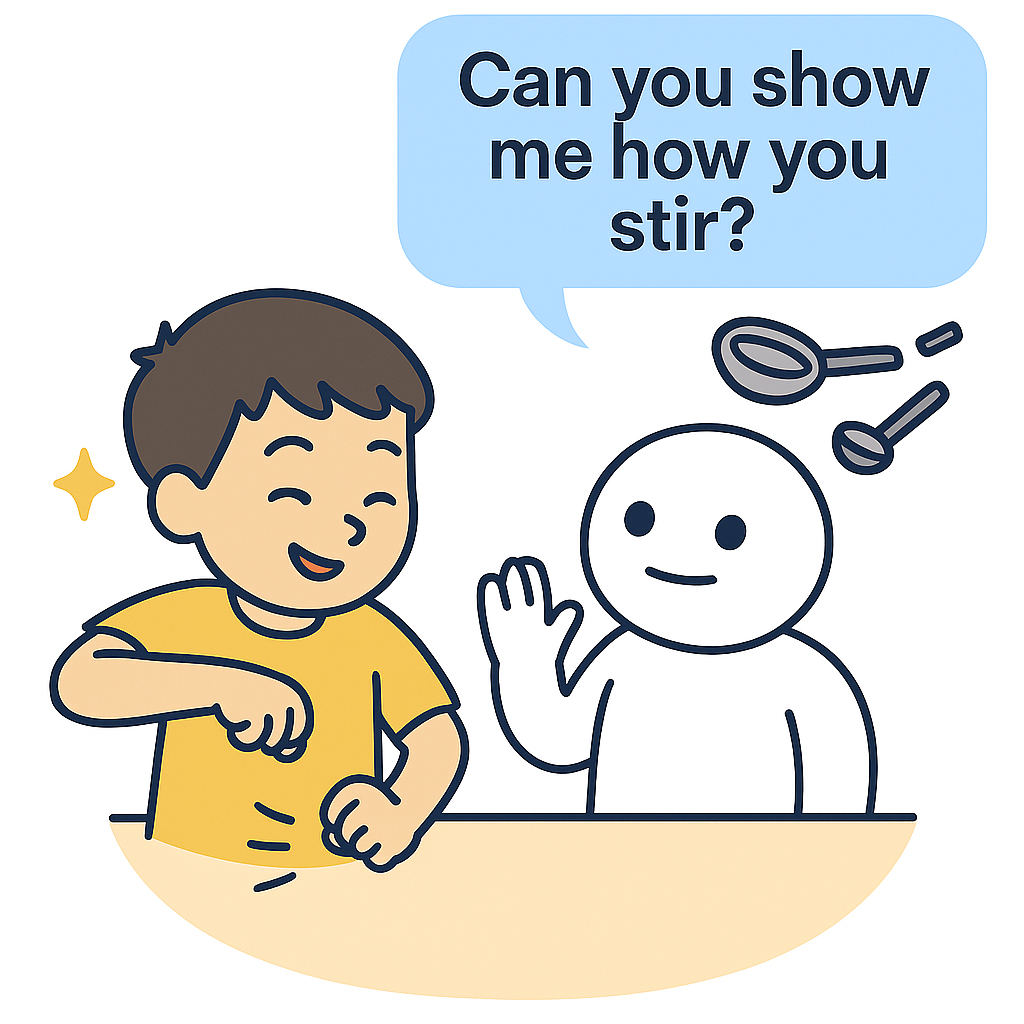}
        \caption{Gesture Imitation}
        \label{activity_gesture}
    \end{subfigure}
    \hfill
    \begin{subfigure}{0.24\textwidth}
        \includegraphics[width=\linewidth]{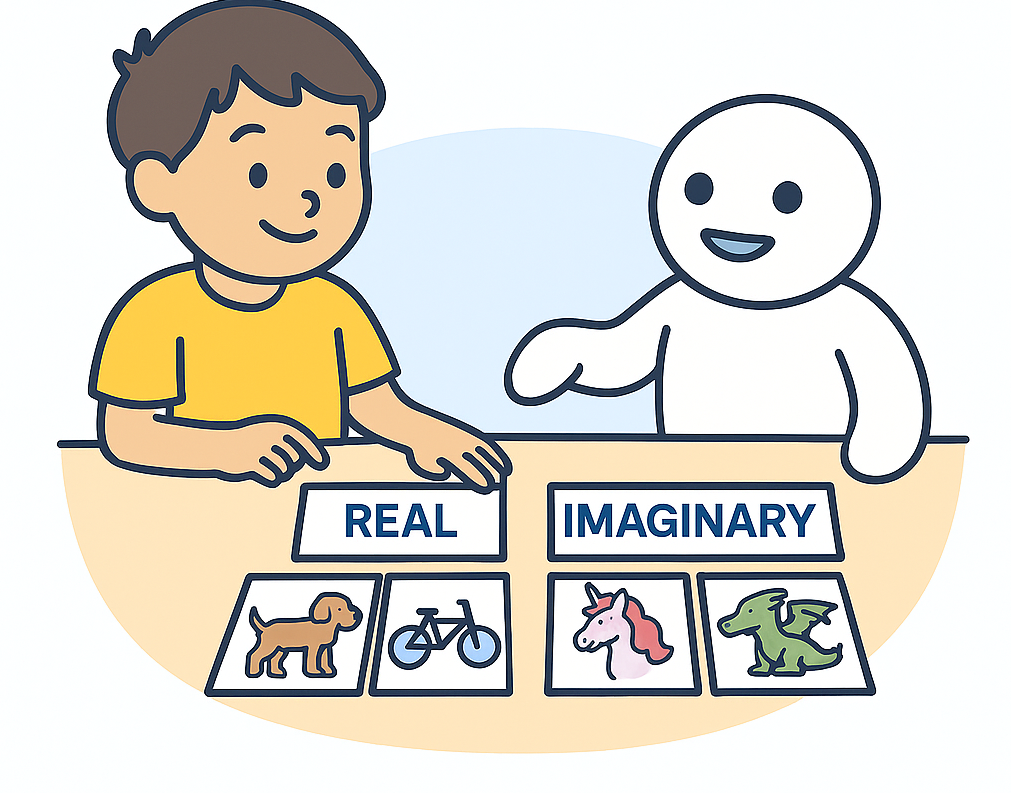}
        \caption{Turn-taking/sorting}
        \label{activity_turn}
    \end{subfigure}
    \hfill
    \begin{subfigure}{0.24\textwidth}
        \includegraphics[width=\linewidth]{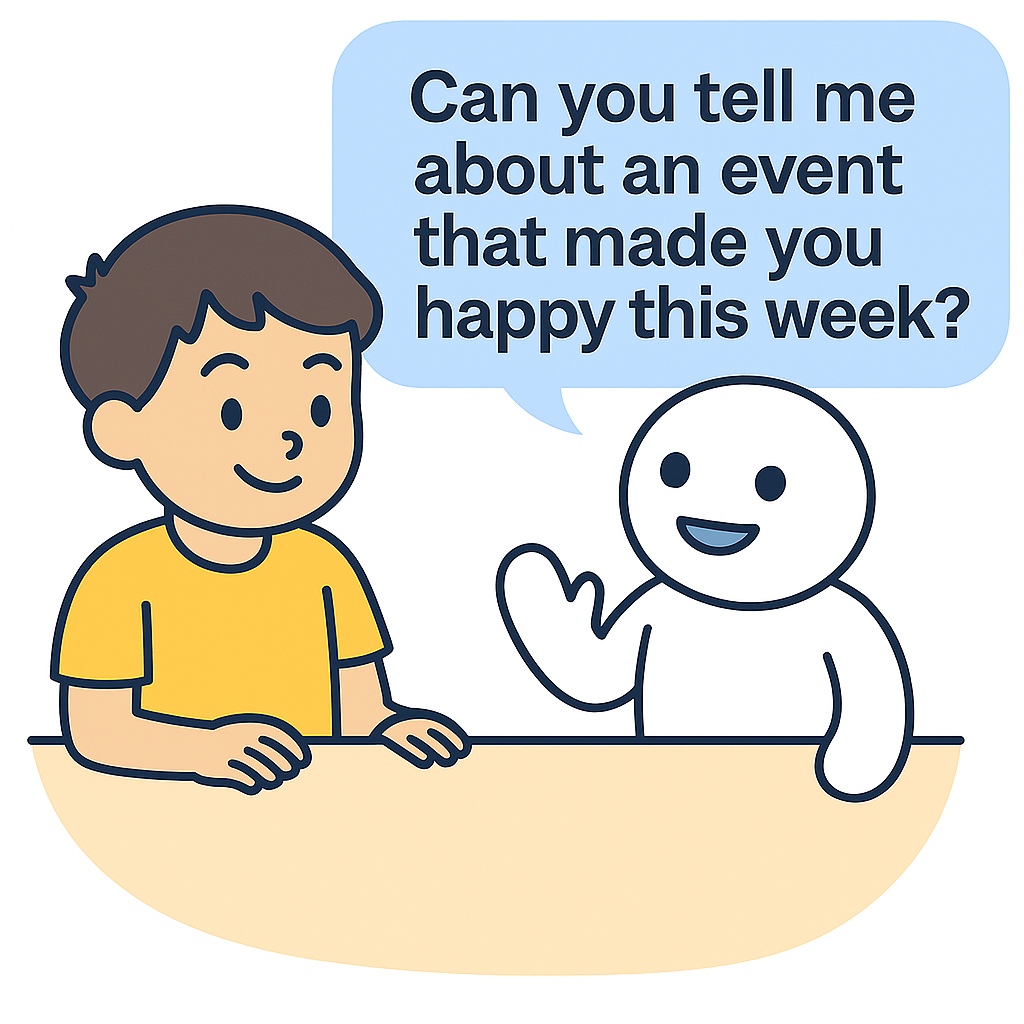}
        \caption{Emotion Expression}
        \label{activity_emotion}
    \end{subfigure}

    \caption{Example activities presented during the focus groups. Visuals taken from the slide deck that was shown to the participants.}
    \label{fig:activity4}
\end{figure*}

\subsubsection{Design of the Activity Framework}
The activity set was designed to provide opportunities for operationalising three identified domains within a child–robot interaction. To capture these domains, five candidate child--robot activities were developed: 
(1) \textbf{Getting to Know You}, (2) \textbf{Structured Storytelling or Picture Description}, (3) \textbf{Gesture Game}, (4) \textbf{Turn-Taking}, and (5) \textbf{Emotion Expression and Understanding} activities.  The first two activities were designed primarily to elicit language and communication behaviours, while the Gesture and Turn-Taking activities targeted social interaction and non-verbal behaviours. The Emotion Expression and Understanding activity focused on children's recognition, interpretation, and discussion of emotions. 


\textbf{Getting to Know You:} The activity was designed as a low-pressure introduction to the robot, allowing children to become familiar with the interaction while providing initial opportunities to observe communication and willingness to engage. This activity was informed by Abbasi et al.~\cite{9900843}, where the robot first introduced itself, used simple social actions such as offering a fist bump, and asked introductory questions to help children become comfortable before completing wellbeing-related tasks. In our material, example activities included the robot introducing itself, sharing a fun fact, asking simple preference questions (e.g., ``What is your favourite game?''), and asking children to describe how their day was using emoji cards.

\textbf{Structured Storytelling or Picture Description:} The activity was designed to elicit language samples while encouraging discussion of social and emotional content. This activity was inspired by two prior approaches: Abbasi et al.~\cite{9900843}, who used a picture task inspired by the Children’s Apperception Test to ask children what they saw and what might have happened before or after a picture, and Bremner et al.~\cite{7451745}, who used a story task based on a picture to elicit interpretations of events, characters' thoughts, and feelings. In our structured storytelling version, children were asked to help the robot reconstruct a story from a series of pictures by describing events, characters' thoughts and feelings, and possible future outcomes (see Figure~\ref{activity_story}). In the picture description version, children were shown a complex scene and asked to describe what was happening, what characters might be feeling, and what they would do in the depicted situation.

\textbf{Gesture Game:} The activity was designed to provide opportunities to observe behaviour without relying heavily on spoken language. This activity was inspired by Bremner et al.~\cite{7451745}, which included a mime task (preparing a meal) that participants acted out everyday actions. In our material, example activities included emotion mimicry, in which children reproduced emotional expressions shown by the robot or on emotion cards; gesture imitation, in which children copied actions such as waving, clapping, or dancing; acting games, in which children acted out everyday activities (see Figure~\ref{activity_gesture}); and action-guessing games, in which the robot attempted to identify the action being performed.

\textbf{Turn-Taking Game:} The activity was intended to elicit social interaction in a playful context, including turn-taking, cooperation, and competition. This activity was inspired by Skantze et al.~\cite{10.1145/2818346.2820749}, who studied turn-taking cues in a collaborative card-sorting game involving two participants in each session (either children or adults) and the Furhat robot. In our material, example activities included category-sorting games, where the robot and child took turns classifying items  (see Figure~\ref{activity_turn}), and matching-pairs games, where both participants alternated turns to find matching cards.

\textbf{Emotion Expression and Understanding:} The activity was designed to explore children's recognition, interpretation, and discussion of emotions. This activity was inspired by Abbasi et al.~\cite{9900843}, where the robot asked children about recent positive and negative memories. In our material, example activities included identifying emotions from facial expressions, matching emotions to social situations, discussing recent emotional experiences, and reflecting on situations in which particular emotions had been felt (see Figure~\ref{activity_emotion}).


Across all activities, the design objective was not to assess wellbeing difficulties directly, but rather to create interaction opportunities through which verbal, behavioural, social, and emotional indicators relevant to wellbeing could be observed in a naturalistic and engaging manner.

\subsection{Focus groups}
\label{sec:focus_groups}

\subsubsection{Focus Group Participants, Recruitment, and Ethics}

\paragraph{\textbf{DLD focus group:}}
Ethical approval for the DLD-focused focus groups was obtained from the Faculty of Education, University of Cambridge. Participants were recruited through purposive sampling using the professional networks of the research team, collaboration with an existing DLD project within the Faculty of Education, and contacts with relevant practitioners and organisations. 
Recruitment materials included an invitation email and participant information sheet describing the study aims, participation requirements, and eligibility criteria. Interested individuals contacted the research team directly, ensuring that participation was voluntary.

Seven participants took part across two DLD-focused focus groups. Participants included two parents of children with Developmental Language Disorder (referred to as the DLD-parents focus group) and five professionals with experience supporting children aged 8--11 years (DLD-professionals focus group). All participants identified as female. The parent participants chose not to report their age. The professional participants ranged in age from 42 to 68 years ((M = 52.4), (SD = 10.8)) and represented a range of professional backgrounds, including family practice, teaching, school-based counsellors, and supervising counsellors.

Participants were selected because of their experience with children and/or DLD, enabling them to provide informed perspectives on the suitability of social-robot interactions for assessing children's wellbeing. Including both parents and professionals allowed the study to capture complementary perspectives on children's communication, social interaction, emotional wellbeing, and support needs.

\paragraph{\textbf{Forced migration focus group:}}

The forced migration-focused focus group was approved by the Swedish Ethical Review Authority. Participants were recruited through purposive snowball sampling. Practitioners with relevant experience working with children from forced migration backgrounds were identified through the research team's professional networks and subsequently invited to recruit additional participants within their own organisations. Recruitment materials, including an invitation email and participant information sheet outlining the study aims, were distributed to potential participants via practitioners. Where possible, participants from the same organisation were grouped together to facilitate more contextually grounded discussion. In cases where a practitioner wished to participate but was unable to recruit colleagues, they were grouped with participants from other organisations.

The forced migration-focused focus groups consisted of two sessions with a total of seven participants. Six participants identified as female and one identified as male. Three of seven participants provided age information ((M = 45.7), (SD = 5.7)). The first group included four professionals from the same organisation, representing diverse backgrounds in psychology, technology and innovation, and health promotion. The second group consisted of three professionals from different organisations, with backgrounds in psychotherapy and nursing.

Participants were selected based on their professional experience with children from forced migration backgrounds and/or within relevant support organisations. This criterion was intentionally broad to capture both those with direct practice experience and those with relevant organisational or technical expertise, enabling diverse perspectives on social robot interactions for children's wellbeing. Practitioners were recruited rather than parents from forced migration backgrounds, given the sensitive nature of forced migration experiences and ethical considerations.

\subsubsection{Focus Group Procedure}
\paragraph{\textbf{DLD focus group procedure:}}
Prior to participation, all participants received an information sheet and provided informed consent through an online Qualtrics form. Participants also completed a short demographic questionnaire collecting information including age, professional background, and familiarity with robots. The focus groups were conducted online via Microsoft Teams and lasted approximately 90--120 minutes. Two researchers facilitated each session using the slide-based discussion materials. 
Participants received a £15 electronic gift voucher in recognition of their time and expertise. 

\paragraph{\textbf{Forced migration focus group procedure:}}

Prior to the sessions, participants completed and returned a consent form via email. Each session was facilitated by two researchers: one primary facilitator led the discussion in Swedish using slide-based materials, while a second English-speaking researcher was present to address questions. Since participants were spread across Sweden, both focus group sessions were conducted online via Zoom, each lasting approximately 90 minutes. To provide participants with a tangible sense of social robots, a NAO robot was displayed on camera by a researcher during the sessions. The sessions were audio-recorded using a dedicated recording device. At the start of each session, participants were invited to give a brief self-introduction including their professional background. Age information was subsequently collected via follow-up emails. Participants were compensated with a 200 SEK electronic gift voucher for their time and contribution.

\subsubsection{Focus Group Protocol}

The focus groups followed a semi-structured protocol with a slide-based discussion led by two pairs of facilitators (one pair leading the DLD-focused groups and the other leading the forced migration-focused groups; all four facilitators are co-authors). Sessions began with facilitators introducing the session and presenting ground rules, including respecting contributions, maintaining confidentiality within the group and staying on topic. Then, the facilitators introduced the project's aim of using social robots to support the assessment of children's wellbeing, and its focus on children aged 8--11 years, including children with DLD and forced migration backgrounds. The facilitators also provided background information on robots and social robots, including examples of child--robot interaction and the potential role of social robots in assessing wellbeing through verbal and non-verbal behaviours.

Participants were then guided through the proposed child--robot activities. For each activity, the facilitators first provided the purpose of the activity, example robot prompts, and illustrative interaction scenarios. For the warm-up activity, participants discussed what kind of ice-breaker would feel welcoming, how to include children with DLD or other communication needs, and how to help shy children feel comfortable. For the picture-description activity, participants discussed whether the task would elicit language from children aged 8--11 years, what kinds of images should be used, how complex the images should be, and how stories could be made culturally applicable.

For the gesture imitation activity, participants discussed whether non-verbal activities would be engaging for children aged 8--11 years, whether such activities would be suitable for children with DLD, which gesture-based ideas were most appropriate, how long children might remain focused, and what cultural considerations might be relevant. For the turn-taking activity, participants discussed whether the proposed activities would be engaging and suitable for children with DLD, and whether collaborative or competitive games would be more appropriate for assessing wellbeing. For the emotion expression and understanding activities, participants discussed how emotional understanding could be incorporated sensitively and whether emotion-focused discussion would be appropriate for children who may have experienced emotional challenges or trauma.

After the activity-specific discussion, participants reflected on the overall approach, including whether the activities aligned with the study aims of sampling language, social behaviour, emotional state, and other signs of wellbeing; which activities seemed most suitable, age-appropriate, and enjoyable for the target groups; and whether alternative activities should be considered. The discussion then moved to practical implementation questions, including how children should first meet the robot, whether they should receive a video or picture in advance, how long the overall session and individual activities should last, and whether interactions should take place individually or in groups.

Participants were also asked to discuss the robot's appearance and social role, including whether the robot should be framed as a friend, teacher, or helper. The protocol then addressed questionnaire-based wellbeing measures, asking participants whether the proposed questionnaires were appropriate, when they should be completed, who should complete them, whose responses would provide the most reliable ground truth, and whether alternative questionnaires should be considered. Finally, participants discussed ethical concerns, including informed consent and assent, privacy and data security, over-attachment or misunderstanding, cultural sensitivities and biases, self-disclosure, and any additional ethical red flags. Each session concluded with a summary and an opportunity for final reflections.

\subsubsection{Iterative Refinement and Adaptation of Materials}

The focus group materials evolved throughout the study to support discussions relating to the project's two target populations: children with DLD and forced migration backgrounds. While the DLD materials underwent a refinement informed by stakeholder feedback, the materials used in the forced migration focus groups were adapted to reflect population-specific contextual considerations. Together, these processes aimed to improve the relevance, accessibility, and appropriateness of the proposed child--robot interaction framework for both target groups.

\paragraph{\textbf{DLD focus group material refinement:}}
The activity set was refined through an iterative participatory design process involving two focus groups. The first focus group (DLD-Parents) was conducted using an initial set of candidate activities that included both structured storytelling and picture-description tasks for language sampling, as well as multiple gesture-based interaction formats such as acting and action-guessing games. Participants were invited to discuss the appropriateness, inclusiveness, cultural applicability, and engagement potential of each activity.

Following the first focus-group feedback, the research team revised the materials before conducting the next focus group. Several activities were simplified and consolidated. Structured storytelling was removed as a separate activity, and picture description was retained as the primary language-sampling task. Gesture-based activities were narrowed to gesture imitation. 
Activity discussion durations were also revised, increasing the proposed discussion time for each activity. In addition, following a review of candidate wellbeing measures, the wellbeing questionnaire component was updated from KINDL~\cite{ravens1998assessing} and SDQ~\cite{goodman1997strengths} to KIDSCREEN-52~\cite{ravens2005kidscreen}, as KIDSCREEN-52 provided broader coverage of the project's multidimensional wellbeing model and was better suited to the exploratory nature of the planned child--robot interaction studies.

The second focus group (DLD-Professionals), therefore, provided an opportunity to evaluate the revised activity set and gather further feedback on its feasibility and acceptability. This iterative process allowed the activities to evolve through a combination of interdisciplinary expertise, evidence from prior literature, and stakeholder perspectives.

\paragraph{\textbf{Forced migration focus group material refinement:}}

The focus group materials were first reviewed and refined within the forced migration research sub-team, then further developed through an iterative participatory design process involving focus group participants. A summary of the conceptual framework underpinning the study \cite{microwellbeing}, comprising five interrelated dimensions of children's wellbeing (physical, social, academic, agency, and emotional), was included as supplementary background information to support participants' understanding of wellbeing as conceptualised in this project.

\begin{figure}[t!]
    \centering
    \begin{subfigure}{0.23\textwidth}
        \includegraphics[width=\linewidth]{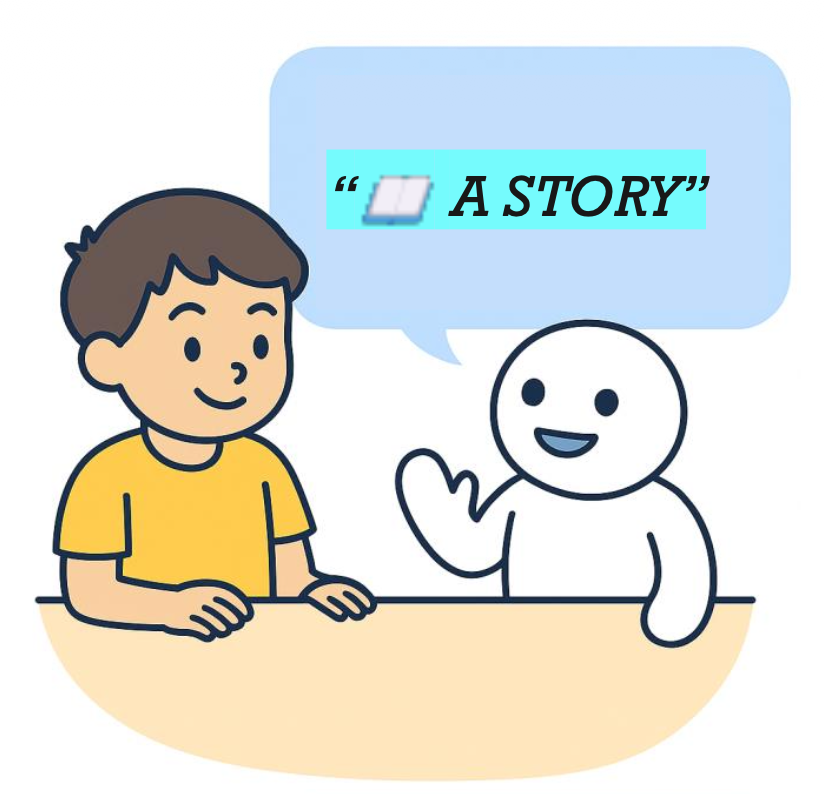}
    \end{subfigure}
    \begin{subfigure}{0.23\textwidth}
        \includegraphics[width=\linewidth]{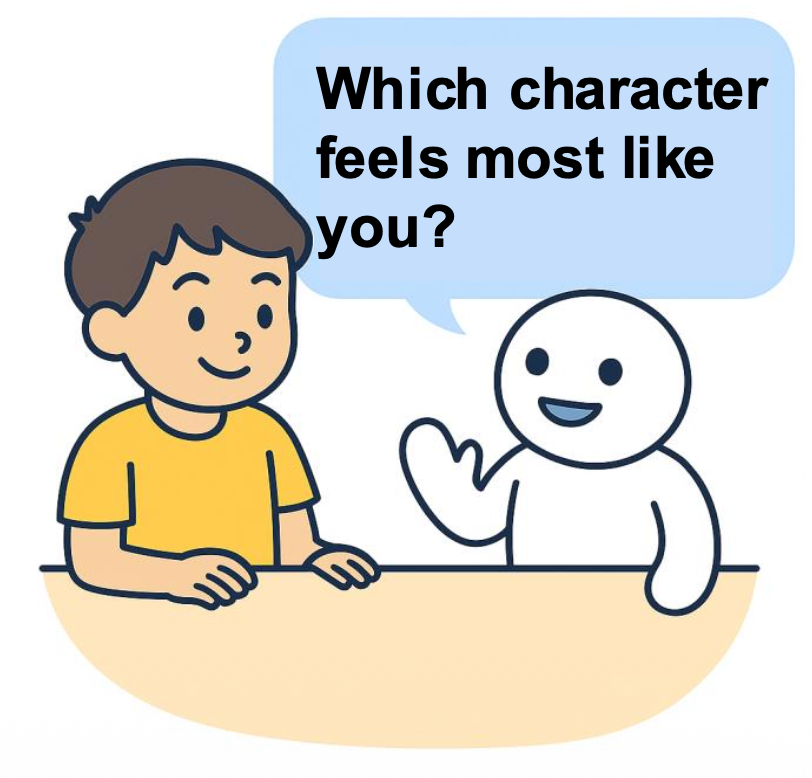}
    \end{subfigure}
    \caption{Story-based Self-disclosure Task. Visuals taken from the slide deck that was shown to the participants.}
    \label{fig:activity_storydisclosure}
\end{figure}

An additional candidate child-robot interaction activity, the \textbf{Story-based Self-disclosure Task}, was also developed during this phase and included in the forced migration-focused group sessions only. This narrative-based reflection task was designed to elicit self-disclosure and resonance through scenario design grounded in validated questionnaires, offering a more engaging and playful alternative that captures richer reasoning and non-verbal insights. This approach was inspired by the Pictorial Scale of Perceived Competence and Social Acceptance for Young Children \cite{pictorial_scale}, in which children are presented with contrasting descriptions of two peers and asked to indicate which one they resemble more, thereby enabling self-relevant judgments without relying on direct verbal questioning. This was adapted to a narrative format using robot-led, third-person storytelling, which prior HRI research has used to elicit listener identification with story characters \cite{storytellerempathey}.
The robot tells a story in which several characters each represent different experiences aligned with one or more constructs from established wellbeing questionnaires, such as the KIDSCREEN-52 \cite{ravens2005kidscreen}, translating questionnaire items into character experiences. The child is then invited to reflect on which character feels most like them and why (see Figure~\ref{fig:activity_storydisclosure}).

Similar to other candidate activities, the facilitator materials include an overview of the task purpose, example robot prompts, and discussion points covering potential challenges, enhancements, and how to design emotionally neutral and culturally relatable characters and scenarios. This task is positioned as a proof-of-concept, with future iterations and empirical testing needed to evaluate its reliability and validity. The narrative can be co-created with children to ensure cultural relevance and authenticity, making it particularly suitable for adaptation with children from forced migration backgrounds.

The candidate activities were presented in the following order in the first focus group: (1) Getting to Know You, (2) Structured Storytelling or Picture Description,
(3) Emotion Expression and Understanding, (4) Story-based Self-disclosure Task, (5) Gesture Game, and (6) Turn-Taking. Activities were ordered from more to less discussion-intensive, allowing richer reflection early in the session when participants were most engaged, with activities (2) and (3) also serving as scaffolding for (4).

Following feedback from the first focus group, two key revisions were made before the second session. First, the activity order was adjusted, with the Gesture Game moved earlier to signal that the activity set extended beyond language-focused interactions. Second, more time was allocated for activity design discussion, as the first group spontaneously raised ethical considerations during activity discussions, indicating the need for more structured discussion in this area. Relevant insights from the first focus group were shared organically during the second session where applicable. The second focus group provided an opportunity to evaluate these revisions and gather further feedback from participants with different professional backgrounds.

\subsection{Data Analysis Procedure}
\label{sec:data_analysis}
\subsubsection{Thematic Analysis}
The focus groups were analysed using reflexive thematic analysis following the six-phase approach described by Braun and Clarke \cite{braun2006, braunclarke2021}. The six phases include familiarisation with the data, generating initial codes, constructing candidate themes, reviewing themes, defining and naming themes, and producing the report. Analysis was primarily inductive, with codes and themes derived from the content of the discussions rather than imposed by a predetermined coding scheme.

\subsubsection{DLD specific procedure and analysis}
The DLD focus groups consisted of one group of parents and one group of professionals. The focus group discussions were recorded and automatically transcribed using the Microsoft Teams built-in transcription. The data were cleaned and anonymised, identifying information, including participant names and references to specific individuals, organisations, and locations, was removed or replaced, and each contributor was assigned a participant identifier (e.g., P1). 

The two focus groups were convened separately and analysed separately in the initial stages, before being combined at a later stage. 
This staged approach allowed the perspectives of each group to be coded and developed on their own terms before being integrated, ensuring that role-specific considerations were captured before shared and cross-cutting themes across the DLD parent and professional groups were identified.
For each group, transcripts were segmented into individual utterances and transferred to a spreadsheet with one row per utterance. Two researchers then independently coded the data. Independent coding allowed each researcher to engage with the data without being anchored to the other's interpretations, broadening the range of analytic perspectives brought to the material and reducing the risk of premature convergence on a single reading \cite{mcdonald2019}. The resulting code sets were transferred to a shared Miro board, where duplicate and overlapping codes were consolidated. Working collaboratively, the researchers iteratively grouped codes into candidate subthemes and themes through reflexive discussion and refinement. The parent and professional analyses were then brought together, with codes and candidate themes from both groups reviewed jointly to develop the DLD-specific themes and subthemes, retaining the source group as an attribute of each contribution so that role-specific nuances remained visible during interpretation. This collaborative process reflects the principles of reflexive thematic analysis, in which themes are actively constructed through researcher engagement rather than treated as passively emerging from the data, and in which dialogue between coders serves to deepen and enrich interpretation rather than to establish a single correct coding \cite{braunclarke2019, braunclarke2021}.

\subsubsection{Forced migration specific procedure and analysis}

The audio recordings were transcribed using OpenAI's Whisper, hosted on a university cluster for secure processing of sensitive personal data. The transcripts were then translated from Swedish to English by a bilingual researcher, without the use of AI tools, before being cleaned and anonymised prior to analysis. Identifiable information, including names and organisational references, was removed or replaced, and participants were assigned pseudonymous identifiers (e.g., P01).

Data from both focus groups were analysed together as a single dataset, as both groups comprised practitioners with relevant professional experience, enabling the identification of patterns across the sample. Similar to the DLD group, transcripts were segmented into individual utterances and entered into a spreadsheet for coding. Two researchers independently coded the data, drawing on backgrounds in human-robot interaction and public health respectively. This dual-researcher approach foregrounded subjectivity and reflexivity as central to the analytic process, in line with reflexive thematic analysis \cite{braunclarke2021}. Codes were then reviewed and iteratively refined through discussion. Through collective review of all codes, emerging subthemes were identified using post-it notes and progressively consolidated into broader themes. Initial themes and subthemes were subsequently transferred to a shared Miro board in preparation for cross-group analysis with the DLD dataset (see Figure \ref{fig:themes_migration} in Appendix \ref{miro_appendix}).

\subsubsection{Combined analysis}

Following the DLD-specific and forced migration analyses, three researchers conducted a cross-analysis synthesis to identify overarching themes. The themes and subthemes generated from the DLD parent, DLD professional, and forced migration analyses were transferred to a new Miro board and reviewed collaboratively. Through iterative discussion, researchers compared how considerations from each analysis related to one another, regrouped related subthemes, and refined their scope and definitions. 
This process focused on developing higher-level themes that synthesised the thematic structures generated through the population-specific analyses. 
As substantial conceptual overlap was observed between the DLD and forced migration themes, only the resulting overarching themes are reported. Population-specific subthemes are instead presented in the DLD and forced migration results sections and referred to as population-specific contributions to overarching themes.

\begin{figure*}[t!]
    \centering
    \includegraphics[width=0.9\linewidth]{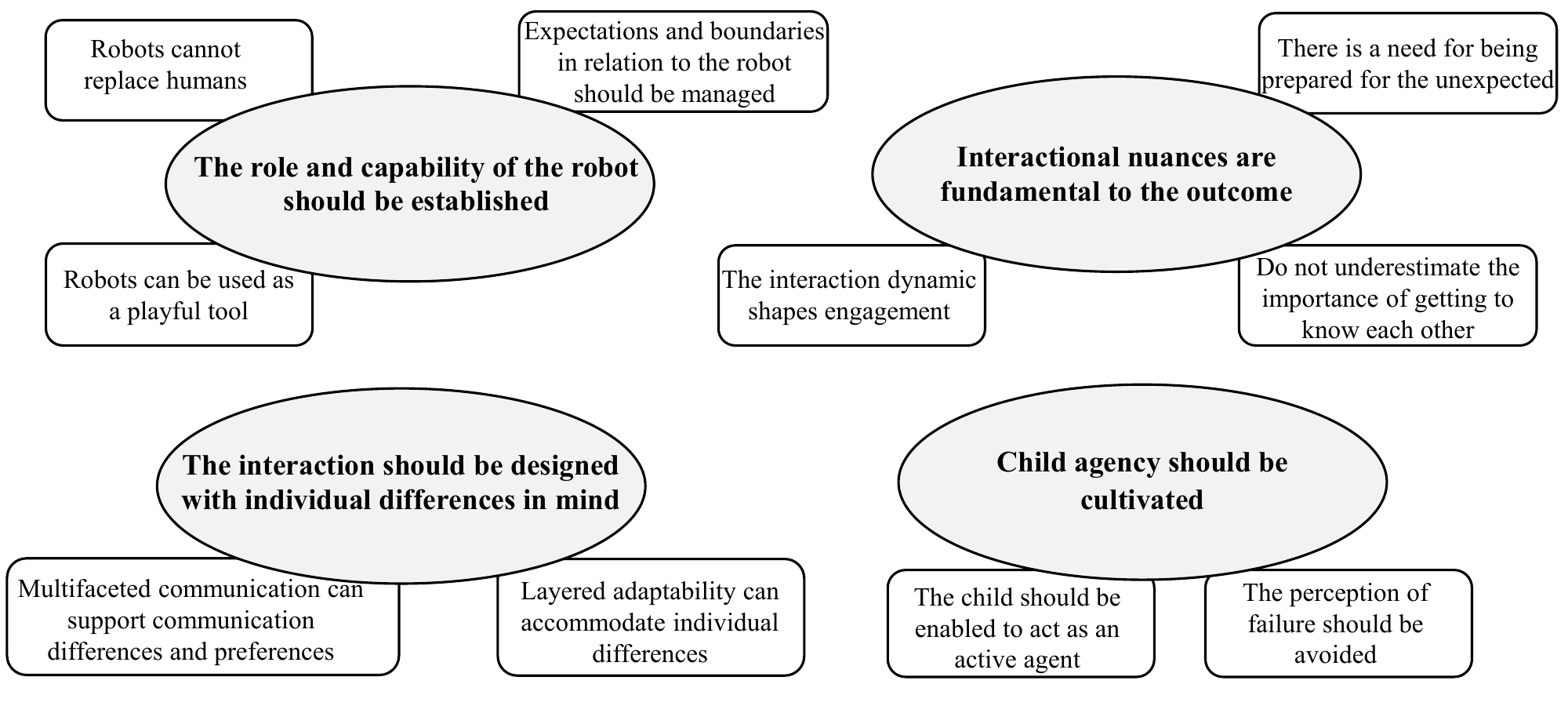}
    \caption{Overarching themes obtained from DLD and forced migration focus groups.}
    \label{fig:themes}
\end{figure*}

\section{Results} 
In this section, we present the findings from the thematic analysis. We first report the overarching results identified through the combined analysis of the DLD and forced migration focus groups in Section~\ref{results_overarching}. We then examine how these themes were reflected in the perspectives of participants from the DLD and forced migration focus groups in Section~\ref{results_DLD} and Section~\ref{results_migration}, respectively.

\subsection{Overarching Results}
\label{results_overarching}
The combined analysis from both DLD and forced migration focus groups resulted in four themes: (i) The role and capability of the robot should be established; (ii) Interactional nuances are fundamental to the outcome; (iii) The interaction should be designed with individual differences in mind; and (iv) Child agency should be cultivated. Figure~\ref{fig:themes} provides an overview of the themes and associated subthemes.

\subsubsection{The role and capability of the robot should be established}
Before addressing child-robot interaction design specifically, participants emphasized the importance of first establishing a clear understanding of the robot's role and capabilities. This includes how the robot is positioned as a playful tool for children, the need to manage expectations and boundaries, and how the robot should complement rather than replace human within the broader healthcare system. Three subthemes emerged under this overarching theme: \textit{Robots can be used as a playful tool}, \textit{Expectations and boundaries in relation to the robot should be managed} and \textit{Robots cannot replace human}.

\paragraph{\textbf{Robots can be used as a playful tool:}}
Participants recognized the strength of using a robot in wellbeing scenarios, noting that its inherently fun and playful nature makes it particularly well-suited for this context. Children may feel more comfortable talking to a robot, especially when topics relate to shame, as it could help reduce embarrassment. Although robots come in different forms and sizes, participants noted that children may find different appeal depending on the size. Smaller robots may feel cute and approachable, while larger ones may feel more exciting.

\begin{quote}
    ``\textit{...I think the child will probably find it quite fun and exciting. It creates a bit of distance, so it doesn’t feel as serious. It’s very similar to the idea behind using dolls or other toys. To help the child shift their focus away from the part that feels serious. I think the children, most of them, will find it enjoyable.}'' -- Participant 1, Forced Migration Focus Group
\end{quote}

Participants also discussed how to preserve this playful impression throughout the interaction to sustain children's active participation. Interactivity was highlighted as key, as tasks should be made playful rather than simply asking questions directly. 

\begin{quote}
    ``\textit{If they’re going to answer questions, for example, they might stand on the floor and make their rating using different sizes of stuffed animals. So, it can be playful... And that usually gets the children, the children become interested and engaged in what you’re doing.}'' -- Participant 5, Forced Migration Focus Group
\end{quote}

This could be achieved through simple, concrete interactions that are low in demand and easy to focus on, while also incorporating humour. Robot humour was seen as both an engagement tool and a social mediator that facilitates the interaction.

\begin{quote}
``\textit{Would the robot be able to answer or would the robot be able to say or would you have a pre-script to say no, you know that's too difficult or something like that, but getting a bit of humour in? Because my child finds it difficult to laugh, and to get a child to laugh will open up and instantly relax. The moment he showed humour, that was it. Somehow the understanding became better and everything}'' -- Participant 2, DLD (Parent) Focus Group
\end{quote}

\paragraph{\textbf{Expectations and boundaries in relation to the robot should be managed:}}

Following the previous subtheme, which identified that children largely appreciate interacting with robots, participants raised the considerable importance of managing expectations and boundaries in relation to how children perceive the robot.

A key concern was the blurring of boundaries between the robot and a human. Children tend to build trust with robots relatively easily, which risks attachment, particularly for children who have poor connections with adults. Participants suggested preparing for this by having an adult nearby to gauge and manage the boundary.

\begin{quote}
    ``\textit{They very quickly start asking questions like when is your birthday. And it quickly becomes very human for a child when interacting with a robot. I think you can quite quickly develop a very high level of trust in a robot. And to feel that it is very easy to talk about things. }'' -- Participant 3, Forced Migration Focus Group
\end{quote}

Participants also identified that this natural trust raises privacy concerns. The robot is not simply a toy, interactions would be recorded and would not remain confidential between the child and the robot alone. Participants highlighted two implications of this that should be clearly communicated to children: that the information shared would be seen by others, and that someone would act on it if serious concerns arise. Being transparent about this could help children understand the purpose of the interaction and reduce confusion. At the same time, participants emphasized the importance of data safety, ensuring that information shared is handled responsibly and not beyond its intended use.

\begin{quote}
    ``\textit{What is the purpose of sharing this? The purpose is that someone else should understand the child. The intention is not that you should share your secrets with someone. No, rather that it would be an assessment that then helps with prevention treatment. }'' -- Participant 2, Forced Migration Focus Group
\end{quote}

To help establish clear boundaries, participants emphasized that the robot's purpose should be explicitly clarified, and that it should be introduced neither as a human nor as a pet, to help children form an accurate understanding of what they are interacting with.

Managing children's expectations of the robot's capabilities was also important. Pre-exposure to the robot before the actual session, as well as a clear introduction at the beginning of the interaction, could help align expectations. Without this, children's own perceptions of the robot could lead to mismatched expectations, compromising the robot's predictability as an advantage. This could potentially disrupt the interaction, for example if the robot handles lived experiences or emotions in a way that leaves the child feeling misunderstood. Participants also raised concerns about potential AI bias, as analytical models trained on data from typically developing children may not accurately reflect the experiences of the target population.

\begin{quote}
``\textit{There also may be an element of unconscious bias that needs to be taken into account when programming the robot... My only other concern was that the diversity within DLD is quite broad, so will it be a 'fit all' robot or only certain `levels' of DLD}'' - Participant 2, DLD Focus (Parent) Group
\end{quote}

\paragraph{\textbf{Robots cannot replace humans:}}
Moving beyond the interaction itself, participants further reflected on the role of the robot within the broader healthcare system. They recognized the potential of involving robots in the screening process to improve accessibility and efficiency, and even envisioned roles for the robot beyond wellbeing assessment, such as supporting children during other clinical procedures. However, participants were clear that in any case the robot should serve as a supplementary tool, a more playful one, rather than a replacement for human involvement. 

\begin{quote}
    ``\textit{If I were the child in that kind of situation, I would most of all want to talk to a robot. And absolutely not with any boring adults, that’s my spontaneous thought. But it can vary. But I liked what you said about offering possibilities. Do you want to talk to this kind nurse or kind doctor, or do you want to play with the robot?}'' -- Participant 4, Forced Migration Focus Group
\end{quote}

Participants also highlighted reasons why the robot should not replace human involvement. Robots were seen as less suitable for handling sensitive subjects, where human judgment and empathy are essential. Furthermore, failing to communicate the robot's role clearly could leave users feeling devalued and damage their trust in the system.

\begin{quote}
``\textit{A sensitive approach that needs a person that needs empathy, it needs pace, it needs breaks and time... This falls into anything that a robot could appropriately manage just by asking how do you feel and when did you last feel like that? Though even those two questions are so whole. That actually I'm just, I'm going to just say that I'm sorry. No, I don't. I don't think that this is for that. That's just me.}'' -- Participant 4, DLD (Parent) Focus Group
\end{quote}

\subsubsection{Interactional nuances are fundamental to the outcome}
Beyond establishing the robot's role and capabilities, participants pointed out that the success of robot-mediated wellbeing assessment depends on the finer, moment-to-moment qualities of the interaction. Across the discussions, they drew attention to how familiarity is built at the outset, how the dynamics of the exchange sustain or undermine engagement, and how unanticipated moments are accommodated. This overarching theme comprised three subthemes: \textit{Do not underestimate the importance of getting to know each other}, \textit{The interaction dynamic shapes engagement}, and \textit{There is a need for being prepared for the unexpected}.

\textbf{Do not underestimate the importance of getting to know each other:} 
Participants emphasised that meaningful interaction depends on an initial period of warming up, during which the child becomes familiar with the robot before any substantive task begins. Professionals supporting children with DLD described this process as highly individual as children settle at different rates, and an effective opening adapts to the particular child rather than following a fixed script.

\begin{quote}
``\textit{...when you're talking about icebreakers, to be honest with you, what I often find is the best icebreaker is whatever is organic between me and the young person... each young person is different. So it's working with them.}''
– Participant 5, DLD (Professional) Focus Group
\end{quote}

The same professionals proposed managing this phase through advance preparation. A short introductory video, presented from the robot's perspective, could familiarise the child with the robot and convey basic expectations before the session and hence reducing the likelihood that the encounter is approached as an unstructured novelty.

\begin{quote}
``\textit{I think it might be useful to provide some kind of video in advance... even if it's from the robot's perspective, hi, my name is so and so... some kind of ground rule there... so that information is already there and they're prepared... They're not gonna walk into the room and it'd be like Christmas with a new toy.}'' – Participant 3, DLD (Professional) Focus Group
\end{quote}

In forced migration contexts, participants similarly described an initial handover, in which an adult introduces the robot before responsibility for the interaction is passed to it, and noted that this transition itself needs to be designed carefully. The unfamiliarity that requires such management was also viewed as an asset: participants observed that the robot's novelty initially prompts caution but quickly gives way to curiosity and engagement.

\begin{quote}
``\textit{there it was like they wanted to approach it carefully, and then it was super exciting when something happened. That was exactly the novelty effect.}'' – Participant 7, Forced Migration Focus Group
\end{quote}

\textbf{The interaction dynamic shapes engagement:} 
This subtheme drew a wide range of contributions, and participants converged on the view that engagement is shaped less by the activities themselves than by how the exchange unfolds between child and robot. A recurring concern, particularly among professionals supporting children with forced migration backgrounds, was that the proposed activities were largely robot-led, positioning the child as a respondent rather than an initiator. Participants argued that engaging interaction requires reciprocity, with scope for the child to introduce ideas and shape the exchange. This was closely tied to the robot's responsiveness: contingent feedback that accurately reflects what the child has expressed was seen as essential for sustaining the conversation and for establishing trust, like the clinical practice of reflecting a child's contributions to convey not only comprehension but the correct nuance.

\begin{quote}
``\textit{In all these different scenarios, a lot of it is based on the robot leading the conversation and observing, and then the child responding in some way. What capacity will this robot have for it to be the other way around? Rather that the child initiates different things, comes up with ideas, and wants to try things together with the robot.}'' – Participant 3, Forced Migration Focus Group
\end{quote}

Parents of children with DLD focused on the robot's affective presentation and noted that children may attend closely to the robot's facial expression and body language and may misread an unintended or flat expression, and therefore argued that the robot should convey consistently positive affect. They further cautioned against concluding the interaction on an emotionally heavy note, suggesting that a light closing activity could help the child leave with a positive impression.

\begin{quote}
``\textit{My son, because he spoke so late, focuses on body language. So the amount of times he would come to me and said... Why are you angry? And I was like, I'm not angry... So I think the way the robot would look would also have a massive impact in how the child perceives the robot. It needs to be just positive all over.}'' – Participant 2, DLD (Parent) Focus Group
\end{quote}

Engagement was also linked to the structure and pacing of the interaction. Clear rules and a well-defined task structure were considered supportive, and parents noted that some children need time to observe before participating rather than being moved directly into activities. Pacing was especially important for professionals working with children with forced migration backgrounds, who emphasised children's limited attention span and the need to segment the session with breaks and changes of activity.

\begin{quote}
``\textit{How long did you estimate it would take? Because they're kids... Half an hour for them is almost three hours.}'' – Participant 6, Forced Migration Focus Group
\end{quote}

Finally, professionals supporting children with DLD highlighted the robot's potential as a social mediator within the interaction. They envisaged the robot modelling behaviour alongside the child, moderating exchanges such as turn-taking when a child interrupts or speaks over it, and offering a more holistic view of the child's communication. The robot was also seen as potentially reducing the pressure children may feel to please an adult, although participants were uncertain how such mediation could be reliably realised.

\begin{quote}
``\textit{...if the child is like interrupting the robot or speaking over the robot. Is there going to be... something that can moderate that... so that it can be kind of interactive if the child decides to... start speaking over it or not taking turns, interrupting. Is there going to be something that it can do in that scenario?}'' – Participant 1, DLD (Professional) Focus Group
\end{quote}

\textbf{There is a need for being prepared for the unexpected:} 
Participants anticipated that robot-mediated assessment would inevitably give rise to unpredictable moments, ranging from interruptions and shifts in attention to emotional distress or unexpected disclosures. Professionals supporting children with forced migration backgrounds argued that the priority should be readiness to respond to such moments as they occur, rather than attempting to predict them in advance.

\begin{quote}
``\textit{Maybe it's more about having a safety routine. In one way or another. Being prepared to pick up on these children's reactions rather than trying to predict them. Because it could be anything.}'' – Participant 7, Forced Migration Focus Group
\end{quote}

Professionals supporting children with DLD focused on safeguarding in particular, noting that activities designed to elicit how children feel may surface disclosures that require a clear protocol and a designated person able to act on them.

\begin{quote}
``\textit{...this opens up the opportunity for any safeguarding disclosure... if a child discloses something that's kind of concerning around safeguarding, what's the protocol and making sure that there is someone in the room that... can follow up on that... making sure that that's absolutely paramount first.}'' – Participant 3, DLD (Professional) Focus Group
\end{quote}

The mechanism proposed across both contexts was the presence of a human support person during the interaction. Beyond following up on disclosures, this person could intervene if a child became distressed and could cue the robot to wind down by reading the non-verbal signals that a child has had enough. Participants supporting children with forced migration backgrounds likewise stressed ensuring that adults are present should a child need support.

\begin{quote}
``\textit{I'm wondering about the person being alongside... Would there be... things that they might say to the robot... [a] cue for the robot to go, do you know what, I think it's maybe time for us to do something else... the role of the person in the room to be able to help the robot to read the cues that we would pick up as human to human interaction.}'' – Participant 4, DLD (Professional) Focus Group
\end{quote}

\subsubsection{The interaction should be designed with individual differences in mind}

Participants positioned individual differences as an important consideration in the design of child-robot interactions. Across the discussions, participants emphasised that children vary considerably in their communication abilities, preferences, developmental characteristics, backgrounds, and lived experiences, making a ``one-size-fits-all'' approach to wellbeing assessment inappropriate. Participants therefore advocated interaction designs that can accommodate diverse needs while remaining accessible to children with differing communication needs and experiences. This overarching theme comprised two subthemes: \textit{Layered adaptability can accommodate individual differences} and \textit{Multifaceted communication can support communication differences and preferences}.


\paragraph{\textbf{Layered adaptability can accommodate individual differences:}}

Participants consistently emphasised that a ``one-size-fits-all'' approach would be unlikely to support meaningful wellbeing assessment across children with diverse communication needs. Participants highlighted substantial variation in children's developmental characteristics, communication styles, interests, cultural backgrounds, emotional expression, and previous experiencess, and consequently advocated a layered approach to adaptation.

\begin{quote}
``\textit{I was thinking that that this picture is maybe a bit babyish for 10 to 11 year olds. So I think you'd have to have a variety of pictures and subjects and maybe before you show the picture, find out if there's a. What the child's interests are. Yeah, so I guess it just needs to be a bit more. Yeah, suited to that particular unique child.}'' -- Participant 3, DLD (Professional) Focus Group
\end{quote}

One way participants proposed achieving such adaptability was by providing children with options and alternative pathways through the interaction. Participants proposed allowing children to choose between different topics, images, and verbal or non-verbal activities, thereby accommodating diverse preferences and strengths while reducing interactional pressure. Such flexibility was considered particularly important when children required support to generate responses, experienced difficulties discussing or expressing emotions, or preferred different ways of engaging with a task.

\begin{quote}
``\textit{On that last one with the guess the action, there could be a possibility of mind going blank as well. So just do something now and the robot will guess like don't know what to do. So if there was a like a bank of images just in case like an ice cream, a swimming pool... Do you want to pick something from here that you could do?'}'' -- Participant 1, DLD (Parent) Focus Group \end{quote}

Finally, participants stressed that adaptation should also account for children's previous experiences and potential sensitivities. They highlighted that children's previous experiences may shape how they respond to particular activities, images, or discussion topics. In particular, participants emphasised the importance of avoiding materials that could inadvertently trigger distressing memories or place undue emotional demands on children.

\begin{quote}
``\textit{It [the robot] has a bit of a Star Wars vibe and kind of storm trooper like elements, and could be similar to a soldier... So just keeping in mind that children with war experiences might react negatively to that...You can’t avoid everything. But military and police themes should definitely be avoided. }'' -- Participant 1, Forced Migration Focus Group
\end{quote}

Taken together, participants advocated a form of layered adaptability in which content, pacing, complexity, and interaction pathways could be adjusted to accommodate children's diverse communication needs, preferences, and experiences.

\paragraph{\textbf{Multifaceted communication can support communication differences and preferences:}}

Participants consistently emphasised the value of multifaceted communication approaches. They highlighted considerable variation in children's communication abilities, preferences, and emotional expression, and argued that robot-mediated wellbeing assessment can be supported by multimodal interaction strategies, including gestures, images, emojis, music, and other visual supports. Such approaches were viewed as particularly important for children who experience difficulties expressing themselves verbally, prefer visual forms of communication, or require reduced linguistic demands during interaction.

\begin{quote}
``\textit{I would say that, you know, in the room I'm talking to children. But when we talk about emotions, I'll use colours and I'll use, you know, emojis and a whole lot of different ways for them to actually express an emotion.}'' -- Participant 1, DLD (Professional) Focus Group
\end{quote}

\begin{quote}
``\textit{If you build a world or use blocks, something else gets activated beyond just the linguistic part. }'' – Participant 2, Forced Migration Focus Group
\end{quote}

Participants further noted that some forms of emotional communication may be more accessible than others. In particular, recognising or discussing emotions through external examples, images, or characters was often perceived as easier than directly expressing one's own emotional experiences. As a result, participants suggested that activities could scaffold emotional communication by beginning with emotion recognition before moving towards more personal forms of expression.

\begin{quote}
``\textit{I'm thinking of some neurodivergent young people that I work with, where they may be able to recognise on a card or on someone else what an emotion is, but expressing them themselves. Very, very difficult.}'' -- Participant 5, DLD (Professional) Focus Group
\end{quote}

Finally, participants highlighted that communication may be supported through the involvement of an adult who can facilitate children's engagement with activities and support interaction with the robot. Participants also noted that communication may differ when children interact with a robot compared with a person, creating uncertainty about how children interpret and respond to robot-mediated interactions. As a result, several participants emphasised the potential value of adult support, particularly when children require assistance engaging with activities or when interactions involve potentially overwhelming experiences.

\begin{quote}
``\textit{if you like particular children who've experienced trauma, who may be refugees, that in the instance with the testing and building the concept, there has to be therapeutic support because none of us really know what the impact of that could be in terms of speaking about things, because we already recognise that, you know, children will probably know the difference between a robot and a person. Or will they? And you know, what difference does that make?}'' -- Participant 1, DLD (Professional) Focus Group
\end{quote}

Taken together, participants advocated a multifaceted approach to communication that combines verbal and non-verbal expression, multimodal support, and adult facilitation to accommodate children's diverse communication needs and preferences.

\subsubsection{Child agency should be cultivated}
Participants’ accounts positioned child agency as an important consideration in the design of child–robot interactions. Across the discussions, participants emphasised the importance of designing interactions that support children’s autonomy, enable opportunities for decision-making, and facilitate engagement without fear of failure or evaluation. This overarching theme comprised two subthemes: \textit{The child should be enabled to act as an active agent} and \textit{The perception of failure should be avoided}.

\paragraph{\textbf{The child should be enabled to act as an active agent:}}

Participants highlighted the importance of creating opportunities for children to take initiative during the interaction. Reflections centred on the power dynamics between the child and the robot, particularly in relation to decision-making. Participants advocated for interactions in which children were able to make decisions on behalf of, or in relation to, the robot. This was contrasted with child–adult interactions, where adults are often assumed to hold decision-making authority. The robot was therefore perceived as offering a context in which children might more freely explore and express their own agency.
\begin{quote}
``\textit{…the child gets to decide over the robot, and in that way, you can see how comfortable the child is making decisions about different things, things we wouldn’t do in a relationship with an adult.}'' -- Participant 1, Forced Migration Focus Group
\end{quote}

Participants also discussed the value of positioning the robot in a role where it could receive support from the child. Allowing the child to help or guide the robot was perceived as another way of fostering agency and encouraging active participation.

\begin{quote}
``\textit{I like the idea of the robot needs help, so then the child is there to sort of save the day... I could see him being happy with that kind of scenario ... because he's quite drawn to younger children and I think he gets that sense of being an expert for once.''} -- 
Participant 1, DLD (Parent) Focus Group 
\end{quote}

\paragraph{\textbf{The perception of failure should be avoided:}}

Participants expressed a belief that children often strive to “do well” in interactions with adults and may attempt to identify the “correct” response in order to satisfy perceived expectations. While participants were uncertain about how strongly this tendency would transfer to child–robot interactions, they emphasised the importance of considering social desirability and performance pressure in the interaction design.
\begin{quote}
``\textit{Understanding what an adult is after can often be a big challenge. Children want to please adults; they search for `What is the adult expecting me to say or should say?' And then they look for the right answer. So that’s something to think about, how to structure it in a good way.}'' -- Participant 7, Forced Migration Focus Group
\end{quote}

Participants further noted that the perception of failure might not only relate to the robot itself but also to the adults facilitating the interaction and, potentially, to parents who may later be informed about the child’s responses or behaviour.

To reduce pressure and minimise fear of failure, participants suggested explicitly clarifying that there are “no right or wrong answers” during the interaction. At the same time, they problematised the use of excessive positive reinforcement, arguing that overly frequent praise could be perceived by children as insincere or artificial.

\begin{quote}
``\textit{My son is kind of doesn't like compliments too many because he kind of knows `I'm different. I know I can't do it as well as the others. So don't give me a compliment because I know I'm not as good as my brother and my sister.'} -- Participant 2, DLD (Parent) Focus Group
\end{quote}

\begin{table*}[t!]
\centering
\footnotesize
\caption{Themes, subthemes, and population-specific contributions from the DLD and forced migration focus groups.}
\begin{tabular}{p{0.8cm} p{3cm} p{5.3cm} p{4.5cm}}
\hline
\textbf{Theme} & \textbf{Subtheme} & \textbf{DLD Focus Groups} & \textbf{Forced Migration Focus Groups} \\
\hline

\multirow{3}{*}{\rotatebox[origin=c]{90}{%
\parbox{3.5cm}{\centering The role and capability of \\the robot should be established}}}
& Robots cannot replace humans
& Robot is not a fix (professionals)
& Robots as a supplementary tool within the healthcare system \\

\\[-2pt]
\cdashline{2-4}
\\[-2pt]

& Robots can be used as a playful tool
& Robot as a playful tool (parents); Give room for humour and imagination (parents); Low demand in tasks and interaction (parents)
& Incorporating fun and playful elements; Ensuring child comfort \\

\\[-2pt]
\cdashline{2-4}
\\[-2pt]

& Expectations and boundaries in relation to the robot should be managed
& Mitigating ethical concerns (parents); Well-defined robot framing (professionals); Preventing over-attachment and exploitation (professionals); Identifying and differentiating expectations (professionals)
& Clarifying purpose of interaction; Understanding boundaries \\

\\[-2pt]
\hline\hline
\\[-2pt]

\multirow{4}{*}{\rotatebox[origin=c]{90}{%
\parbox{4cm}{\centering The interactional nuances are\\ fundamental to the outcome}}}
& Do not underestimate the importance of getting to know each other
& Making time to warm up (professionals); Overcoming novelty effect (professionals)
& Novelty effect as a positive influence \\
\\[-2pt]
\cdashline{2-4}
\\[-2pt]

& There is a need to be prepared for the unexpected
& Utilise proactive handling mechanisms (professionals)
& Importance of responding in the moment \\

\\[-2pt]
\cdashline{2-4}
\\[-2pt]

& The interaction dynamic shapes engagement
& Appropriate affective robot behaviour (parents); Importance of time and pacing (parents); Structuring well-defined interaction (parents); Create robot capabilities for social mediation (professionals); Consider human relational disruptions (professionals)
& Need for two-way interaction; Accommodating children's attention span \\

\\[-2pt]
\hline\hline
\\[-2pt]

\multirow{2}{*}{\rotatebox[origin=c]{90}{%
\parbox{2.8cm}{\centering The interaction should be designed with individual differences in mind}}}
& Layered adaptability can accommodate individual differences
& Acknowledging trauma (professionals); Provide options with granularity (both); Enable layered adaptation (both); Understand differences in child disposition (both)
& Relatability of tasks; Adapting to individual differences (e.g., trauma history and culture) \\
\\[-2pt]
\cdashline{2-4}
\\[-2pt]
& Multifaceted communication can support communication differences and preferences
& Rethinking non-verbal tools (parents); Enable reduced verbal demands (parents); Facilitate multimodal communication (parents); Contributions of 3-way interactions (professionals); Recognising emotion easier than expressing (professionals); Participation through non-verbal exploration (professionals)
& Adapting to developmental differences  \\

\\[-2pt]
\hline\hline
\\[-2pt]
\multirow{2}{*}{\rotatebox[origin=c]{90}{%
\parbox{1.8cm}{\centering Child agency should be cultivated}}}
& The child should be enabled to act as an active age
& Child helping robot (parents); Enable child initiative (parents); Cultivate child agency (professionals)
& Power dynamics and agency\\
\\[-2pt]
\cdashline{2-4}
\\[-2pt]
& The perception of failure should be avoided
& Avoid over-praise (parents)
& Social desirability; Pressure to perform well; Children's intention to protect family \\

\\[-2pt]
\hline\hline
\\[-2pt]
\end{tabular}
\label{tab:combined_themes}
\end{table*}

\subsection{Insights from DLD Focus Groups}
\label{results_DLD}

The perspectives of participants supporting children with DLD contributed to all four overarching themes. However, the separate parent and professional analyses revealed different emphases that reflect the distinct roles these stakeholders play in children's lives. While parents primarily discussed interactions through the lens of their children's everyday communication experiences, professionals more often considered how robot-mediated wellbeing assessment could be implemented, facilitated, and safeguarded in practice. These complementary perspectives are summarised in Table~\ref{tab:combined_themes} (middle column).

\subsubsection{Insights from DLD-Parents' Focus Group}


Parents frequently reflected on their experiences of supporting children with DLD in everyday interactions. Within the theme \textit{Role and capability of the robot should be established}, parents most strongly emphasised the subtheme \textit{Having the robot as a playful tool}. They described the robot as a low-pressure interaction partner that could support engagement through humour, playfulness, and imaginative activities. Parents valued interactions that felt simple and enjoyable, emphasising that children with DLD may engage more easily when activities are framed as playful experiences rather than tasks that test their abilities or create a fear of making mistakes.
\begin{quote}
``\textit{
So it needs to be very carefully done that there can be no feeling of `I can't do it because I may get it wrong', which can feature quite heavily. The confidence level is quite low and then they rather say `I don't know' because they can get it wrong. So there must be an element of it's all is fine.}''-- Participant 2, DLD (Parent) Focus Group 
\end{quote}

Within the theme \textit{Interactional nuances are fundamental to the outcome}, parents most strongly highlighted the subtheme \textit{Interaction dynamic shapes engagement}. They emphasised the importance of appropriate affective robot behaviour, clear interaction structure, sufficient time for children to process and respond, and opportunities to engage at their own pace. Participants observed that children with DLD often rely heavily on non-verbal cues such as facial expressions and body language. As a result, they pointed out that ambiguous or unintentionally negative expressions may be interpreted as signs of displeasure or disapproval, thereby undermining engagement.
\begin{quote}
``\textit{
If there would be no reaction from the robot, then it would fall flat and that would not work because then you get the `Oh my God, I did something wrong'. So it's a difficult one as long as the robot can interact correctly, it can be really powerful.''} -- 
Participant 2, DLD (Parent) Focus Group 
\end{quote}

Within the theme \textit{Interaction should be designed with individual differences in mind}, parents contributed strongly to both subthemes. They emphasised that children with DLD vary considerably in their interests, communication abilities, confidence, and preferred ways of engaging with activities. Consequently, participants advocated a layered approach to adaptation that could account for factors such as age, individual skill level, personal interests, and differences in children's dispositions, while also providing options with sufficient granularity to accommodate different support needs. Parents further highlighted the value of reducing verbal demands and supporting communication through multiple modalities, including visual supports, music, and other non-verbal forms of expression. 
These perspectives reflected a recognition that children with DLD may not always communicate their experiences most effectively through direct verbal questioning, and that interactions should provide flexible pathways for expression rather than relying on a single mode of engagement.

\begin{quote}
``\textit{...Direct questions won't work... So going back to what you said with the example activities, what's your favourite game? He'd go, like [name] said, `I don't know'. Even now he's 18, but he's grown up friendless...  `How is your day with pictures?' That would be brilliant.}'' 
-- 
Participant 2, DLD (Parent) Focus Group 
\end{quote}

Finally, within the theme \textit{Child agency should be cultivated}, parents highlighted both subthemes. They valued interaction designs that enabled children to take active roles. In particular, participants described opportunities for children to help or guide the robot, as such interactions allowed children to feel competent, remain in control of the interaction, and experience themselves as helping rather than being helped. Parents also cautioned against excessive praise or encouragement, noting that children with DLD may perceive such responses as insincere or disconnected from their own perception of their abilities.
\begin{quote}
``\textit{I think my son would open up if a robot would be there not doing anything, maybe a couple of moves and the initiative comes from my son. So not from the robot, but from my son, because then he leads where it needs to go. I think he's been told for his life so many times what to do and how to do it. He's a bit tired of that.}'' -- Participant 2, DLD (Parent) Focus Group 
\end{quote}

\subsubsection{Insights from DLD-Professional Focus Group}


Professionals drew on experiences of supporting children with DLD across educational, therapeutic, and clinical contexts, leading them to emphasise communication support, adult facilitation, and ethical considerations. Within the theme \textit{Role and capability of the robot should be established}, professionals most strongly emphasised the subthemes \textit{Robots can not replace humans} and \textit{Expectations and boundaries in relation to the robot should be managed}. Participants stressed that robots should not be viewed as solutions to children's wellbeing needs and raised concerns about robots' ability to respond appropriately to sensitive topics or the complexity of children's lived experiences. They advocated clear framing of the robot's capabilities and limitations, cautioning against creating unrealistic expectations about what the robot can understand, interpret, or provide. Professionals also highlighted the importance of preventing over-attachment, particularly for children who may be vulnerable to viewing the robot as a substitute for friendship or social support.

\begin{quote}
``\textit{... a child who feels very alone might feel that they need a friend... So a robot who's almost kind of paired, you know, peer if you like,... might be really helpful to a child. But again, ... I'd be worried about children who might be quite vulnerable or who might be quite `well, they're my friend'. You know, it's like... about kind of exploitation and things like that as well}'' -- Participant 4, DLD (Professional) Focus Group 
\end{quote}

Within the theme \textit{Interactional nuances are fundamental to the outcome}, professionals contributed strongly to all three subthemes. Participants emphasised the importance of allowing time for children to become familiar with the robot and gradually build trust, for example, through opportunities to encounter the robot before the assessment session. They also highlighted the need to account for novelty effects, recognising that initial discomfort, uncertainty, or heightened excitement may influence children's responses and behaviour. Professionals further discussed the value of proactive support mechanisms, including safeguarding procedures for responding to distress or disclosures that may arise during assessment activities. Finally, participants highlighted the robot's potential role as a social mediator that can facilitate conversations and support engagement. At the same time, they recognised that existing relational dynamics with adults may shape how children respond during interactions, including tendencies to provide answers that they believe adults want to hear. Participants suggested that interacting with a robot may reduce some of these perceived expectations and create different opportunities for children to express their views.

\begin{quote}
``\textit{... for the robot, there's maybe them [children] thinking there's no expectations as to what the robot wants them to say, which I think is a really good thing.... If it was me and I was asking a child that might be working really, really hard to answer what they think I might want. Whereas I think the child in these situations might just answer more about what they see and not be so. Maybe aiming to please adults. I think sometimes children do that. }'' -- Participant 4, DLD (Professional) Focus Group 
\end{quote}

Within the theme \textit{Interaction should be designed with individual differences in mind}, professionals contributed strongly to both subthemes. Similar to parents, they emphasised the importance of layered adaptation through flexible levels of support, options with appropriate granularity, and sensitivity to differences in children's dispositions and communication abilities. However, professionals also placed particular emphasis on how children's previous experiences may influence their responses to assessment activities. Participants further noted that recognising emotions or discussing emotions through external examples, such as characters or images, may be easier for some children than expressing their own feelings directly. Finally, professionals highlighted the value of three-way interactions involving the child, robot, and adult facilitator, viewing such arrangements as an important means of facilitating communication, supporting participation, and adapting interactions to individual needs. Participants noted that the presence of an adult could help children engage with activities, recognise when additional time or support is needed, and provide reassurance when interactions involve emotionally sensitive topics or disclosures.
\begin{quote}
``\textit{If it's a child with communication challenges, it might mean they might need a bit longer process and time anyway. But how does the robot interpret that? Because in the room, I think we're as humans, we're probably quite good at it. Well, I'd like to think we're quite good at doing that. }'' -- Participant 4, DLD (Professional) Focus Group 
\end{quote}

Finally, within the theme \textit{Child agency should be cultivated}, professionals primarily emphasised the subtheme \textit{Enable the child to be an active agent}. Similar to parents, they advocated opportunities for choice, participation, and influence over activities and interaction pathways. They also highlighted the importance of keeping the child at the centre of the assessment process rather than positioning them as a passive recipient of assessment.
\begin{quote}
``\textit{All are important, including the robot, but I think that the centre of has to be the child because it's what the child thinks and feels that that is important here.}'' -- Participant 2, DLD (Professional) Focus Group 
\end{quote}

\subsection{Insights from Forced Migration Focus Groups}
\label{results_migration}

Participants supporting children with forced migration backgrounds contributed perspectives across all overarching themes and subthemes, summarised in the rightmost column of Table~\ref{tab:combined_themes}. As both focus groups comprised practitioners with relevant professional experience, their insights are reported collectively rather than by group. The following elaborates on how their perspectives contributed to each theme, including insights specific to this population.

Within the theme \textit{Role and capability of the robot should be established}, participants strongly contributed to all three subthemes. While their perspectives on \textit{Robots can be used as a playful tool} aligned with the general findings, they placed particular emphasis on \textit{Expectations and boundaries in relation to the robot should be managed} and \textit{Robots cannot replace humans}, with insights specific to this population. In particular, data privacy emerged as a critical boundary concern, as children from forced migration backgrounds and their families may be especially cautious about how data flows and is used. This caution arises from their fragile trust in the new society, and the fear that information shared could negatively affect their immigration status. This concern about trust in the system also extended to the role of the robot within the healthcare system. Drawing on their experience with adults from forced migration backgrounds, participants reflected that this population may question whether they are valued equally to native-born patients, for instance feeling offended when seen by a nurse rather than a doctor. This raised the concern that encountering a robot instead of a human could leave them feeling even more devalued, further damaging trust in the healthcare system. At the same time, participants noted that this concern may apply more to parents than to children, as children are likely to find the robot engaging and fun, as discussed in the subtheme \textit{Robots can be used as a playful tool}. Nonetheless, this reinforces the importance of clearly communicating that the robot is a supplementary tool, not a replacement for human care.

\begin{quote}
    ``\textit{An adult might feel like, “Oh, you’re just saving money and sending me to yet another AI chat that’s bad and gives silly answers.” But I’m thinking that for the child, this could be really good, and that they would have a much more play based relationship to the situation compared to an adult.}'' -- Participant 4, Forced Migration Focus Group
\end{quote}

Within the theme \textit{Interactional nuances are fundamental to the outcome}, participants contributed to all three subthemes, with stronger emphasis on \textit{Interaction dynamic shapes engagement} and \textit{There is a need for being prepared for the unexpected}. Participants particularly emphasised the importance of contingent feedback for children from forced migration backgrounds. Accurately reflecting not only what the child has expressed but also the correct nuance was seen as especially critical for this population, given their more fragile trust in the system and greater need to feel safe during the interaction. Furthermore, given the more likely trauma histories of this population, participants particularly highlighted the importance of being prepared to handle unexpected triggers during the interaction, rather than attempting to predict them in advance, as discussed in the overarching theme.

\begin{quote}
    ``\textit{It can get complicated if you don’t get feedback that matches what you said, because how are you then supposed to continue the discussion, how do you move forward, and how do you feel trust to keep the conversation going.}'' -- Participant 2, Forced Migration Focus Group
\end{quote}

Within the theme \textit{Interaction should be designed with individual differences in mind}, participants strongly contributed to both subthemes but placed group-specific emphasis on several considerations under the subtheme \textit{Layered adaptability can accommodate individual differences}. Given the migration histories of this population, culture was most frequently raised in relation to the representation of study materials. As children in this group come from diverse regions and cultural backgrounds, participants suggested ensuring that children feel represented, with skin colour and ethnicity highlighted as a starting point. Beyond visual representation, participants also recommended considering the cultural context of study materials and incorporating elements that are shared across backgrounds. At the same time, they noted the importance of remaining attentive to individual differences, for example that emotions are not universally expressed or perceived in the same way. Participants also highlighted the need for multilingual support, as children in this group may have varying levels of proficiency in the language of the country of settlement, suggesting that robot interactions should not rely solely on verbal communication in a single language. Beyond  cultural and language representation, participants also noted considerations regarding trauma history, that everyday objects could serve as potential triggers, which are individual and unpredictable rather than universal. While the aim is to avoid reminding children of trauma where possible, it is difficult in practice to anticipate all individual triggers. This links to the subtheme \textit{There is a need for being prepared for the unexpected}, where having a safety routine in place is prioritized over attempting to avoid all potential triggers. Nevertheless, certain topics carry a higher risk of distress for children from forced migration backgrounds and should be avoided where possible, such as military and police themes, and sensitive subjects including family and finances.

\begin{quote}
    ``\textit{The grass, someone in our focus group said she had been bitten by a snake when they walked through the forest. So even the grass can become an issue. So, you can’t really say specific things. Memories get triggered. }'' -- Participant 6, Forced Migration Focus Group
\end{quote}

Finally, within the theme \textit{Child agency should be cultivated}, participants strongly contributed to both subthemes with population-specific focus. Within the subtheme \textit{The child should be enabled to act as an active agent}, participants highlighted how experiences of power and control may shape how children with forced migration backgrounds engage in child–robot interactions. Participants reflected on how these children may previously have experienced situations in which other people, institutions, or systems exercised authority and control over them. Consequently, interactions in which the robot is perceived as holding power or demanding compliance could risk evoking stress or traumatic memories. The inclusion of parental perspectives within the subtheme \textit{The perception of failure should be avoided} was primarily derived from discussions concerning children with forced migration backgrounds. Participants reflected on how children may be aware of wider societal narratives surrounding families who have been forcibly displaced, and how this awareness may shape their behaviour during interactions. In particular, it was suggested that children might act in ways that seek to protect or avoid exposing their families to perceived judgment.

\begin{quote}
    ``\textit{One should be aware that criticism of the family can be very difficult for these families who have had to hold together and feel they need to defend themselves against the outside world, and that it isn’t criticism. Things that might be experienced as ‘now my parents are setting boundaries for me, is that right or wrong?’ Children can be quite aware that there is a societal discussion about this, and that it could be negative for their parents.}'' -- Participant 1, Forced Migration Focus Group
\end{quote}

\section{Discussion}
Our findings identify four interrelated considerations for designing robot-mediated wellbeing assessment for children with diverse communication needs: establishing the robot's role and capabilities, accounting for interactional nuances, designing for individual differences, and cultivating child agency. Together, these findings suggest that inclusive robot-mediated assessment depends not only on what activities are used but also on how the robot is framed, how the interaction unfolds, how children are supported, and how safety, adaptation, and agency are built into the assessment process.

\subsection{RQ1: Considerations for children with diverse communication needs}
\label{sec:rq1}

In relation to RQ1, our findings extend prior HRI work on robot-mediated wellbeing assessment by shifting attention from whether social robots can elicit wellbeing-related information~\cite{10.1145/3722123,abbasi2024analysing,9900843} to the design considerations that make such assessment inclusive and appropriate for children with diverse communication needs. The theme \textit{Role and capability of the robot should be established} highlights the importance of positioning the robot within the wellbeing assessment process as a complementary tool rather than a substitute for human support. This theme aligns with broader discussions about transparency and expectation management in HRI~\cite{yadollahi2025expectations, horstmann2020expectations}, while further reinforcing the importance of communicating the robot's capabilities, limitations, and purpose in wellbeing assessment contexts. The theme \textit{Interactional nuances are fundamental to the outcome} further suggests that assessment outcomes cannot be separated from the interactional dynamics through which they are produced. While prior HRI research has examined these dynamics in relation to rapport, trust, and engagement in child--robot interactions~\cite{10.3389/frobt.2019.00054, van2020child}, our analysis suggests that they may also influence wellbeing-related information. Therefore, trust-building, pacing, and affective behaviour should be considered not only as factors that shape the interaction experience but also as factors that may influence the interpretation of assessment outcomes. 

The theme \textit{Interaction should be designed with individual differences in mind} reinforces growing calls within HRI for adaptive and personalised interactions~\cite{andriella2025personalising, ahmad2017adaptive, 10.1145/3776734.3794453, 10.1145/3757279.3785610}. However, our findings suggest that adaptation in wellbeing assessment should extend beyond personal preferences or engagement-related factors and account for differences in communication abilities, support needs, and previous experiences. This highlights a challenge for HRI research: balancing the flexibility to accommodate diverse communication needs and multimodal forms of participation with the consistency of assessment procedures and robots' technical feasibility. Finally, the theme \textit{Child agency should be cultivated} extends growing interest in child-centred and participatory approaches within HRI~\cite{https://doi.org/10.1111/bjep.70078, 10.1145/3555118} by highlighting that agency remains important beyond the design process and throughout interaction. For children who may experience communication barriers, our findings support that inclusive wellbeing assessment should create opportunities for choice and initiative, enabling children to participate as active contributors to the assessment process. Together, these findings suggest that robot-mediated wellbeing assessment should be understood as a relational and contextual process rather than a straightforward data-collection task. 

Finally, these findings support our rationale for including DLD and forced migration as two illustrative contexts of diverse communication needs. Across both contexts, stakeholders identified shared considerations for inclusive robot-mediated wellbeing assessment, including the need to define the robot's role, attend to interactional dynamics, accommodate individual differences, and cultivate child agency. These considerations provide a foundation for broader design principles for children with diverse communication needs.


\subsection{RQ2: Considerations for DLD and forced migration context}

\label{sec:rq2}
In relation to RQ2, our findings demonstrate that although many considerations were shared across the two communities, they manifested in ways that reflected the specific communication-related needs and experiences of each population. This suggests that while common design principles provide a foundation for inclusive robot-mediated wellbeing assessment, the implementation of these principles should remain sensitive to the factors that affect how children communicate, participate, and express their experiences.


For children with DLD, participants' reflections highlighted the importance of designing robot-mediated assessment around children's communication needs. Consistent with prior HRI research on neurodivergent children, stakeholders emphasised adaptation, multimodal communication, and flexible forms of participation~\cite{10.1145/3776734.3794564, 10974006}. Our findings further suggest that, in wellbeing assessment context, design should support playful and low-pressure interactions, appropriate affective robot behaviour, reduced verbal demands, and sufficient time for children to process and respond. These findings indicate that HRI research should consider not only how communication is accommodated, but also how interaction design creates opportunities for children with DLD to participate and express themselves during assessment. Furthermore, the findings highlight the role of adult facilitation and support. Consistent with previous HRI research highlighting the communicative benefits of adult involvement~\cite{10.1145/3757279.3785564}, professionals in our study emphasised the role of adults in facilitating interaction, responding to sensitive topics, and contextualising children's responses. Participants also raised concerns about over-attachment to robots, a particularly important consideration given the elevated risk of loneliness among children with language difficulties~\cite{asher1999loneliness}. Together, these findings suggest that, for children with DLD, robots should be carefully positioned within the assessment process and understood as one component of a broader assessment approach. 

For children from forced migration backgrounds, participants' reflections highlighted considerations shaped by experiences of displacement, cultural and linguistic diversity, and trauma. Participants raised concerns about the robot's role within healthcare systems, particularly around data transparency and data flow. These concerns reflect documented patterns of distrust towards government and institutional systems among forced migration populations~\cite{Yang_Hu_Dautenhahn_2025} and broader challenges in building trust around health data collection in this population~\cite{data_health_migrant}. Consistent with calls in HRI for intentional and transparent communication about data practices when deploying robots in sensitive contexts involving children~\cite{safeguarding_2026}, and evidence that privacy and confidentiality concerns can compromise self-report accuracy, particularly among more vulnerable young people \cite{SONESON20251008oxwell}, these findings suggest that robot-mediated assessment requires clear communication about data use and safeguards, and that robots should be positioned within accountable institutional frameworks rather than deployed as standalone tools. Furthermore, participants also emphasised culturally inclusive design, highlighting the need for assessment materials to reflect children's diverse cultural backgrounds and linguistic diversity. This aligns with calls for culturally responsive and multilingual design in HRI research involving migration-background populations~\cite{participatory_refugee_2026, Tozadore_Kuoppamaki_Guneysu_2023}, and suggests that co-creation with children and communities may be essential to ensure relevance and trust. Finally, participants highlighted that avoiding potentially triggering topics entirely is neither realistic nor necessary, but that safety and appropriate safeguarding must be prioritised. As evidenced by trauma care frameworks such as Teaching Recovery Techniques \cite{TRT_2026}, which support controlled exposure within safeguarding structures, our findings suggest that the priority should be ensuring adequate contextualisation, professional oversight, and adult support, consistent with HRI research emphasising that sensitive disclosures require careful human facilitation~\cite{safeguarding_2026}.

\subsection{RQ3: Inclusive and ethical design recommendations}
Several of the concerns participants raised, that a single ``fit-all'' robot may not serve the breadth of children with DLD, that models trained on typically developing children may misrepresent the target population, that a militarised robot aesthetic could distress children with war experiences, and that a robot perceived as exercising authority could re-evoke coercive experiences, are examples of local instances of structural problems documented across critical AI ethics. We read them here through the lens of epistemic justice. Wellbeing assessment is, at root, a question of whose account of their own inner life is heard and credited. Children with DLD and children with forced migration backgrounds already face a heightened risk of testimonial injustice, since communication barriers can lead adults to discount or misread what they express~\cite{duinmeijer2025language, fazel2018preventive}. Delegating assessment to an automated system risks formalising this disadvantage rather than relieving it. The risk is most acute when AI systems can withhold recognition of personhood altogether, misclassifying disabled users as nonhuman or as noise to be corrected, and treating such misrecognition as a natural limitation rather than a designed exclusion~\cite{nakamura2019algorithms}. 

A parallel concern arises from the cultural narrowness of the systems that often underpin social robots. If for example large language models (LLMs) would be used, they often hold a Western, Educated, Industrialised, Rich, and Democratic (WEIRD) psychology; their responses to psychological measures most closely resemble those of WEIRD respondents and diverge sharply as cultural distance increases, a pattern that larger models and multilingual prompting do not reliably remove~\cite{atari2023which}. Children with forced migration backgrounds sit, by definition, outside this frame, so a system that encodes dominant-culture assumptions about how emotions are expressed and what counts as typical behaviour may systematically misread their wellbeing. Cultural meaning is also carried by the robot's physical body: social robots are not neutral artefacts, and design defaults such as predominantly white surfaces are racially marked, so embodiment choices distribute representational benefits and harms whether or not designers intend them~\cite{sparrow2020robotics}. The participant observation that a robot with a militarised, storm-trooper appearance could distress children with war experiences is a concrete instance of this point. Within HRI specifically, these are matters of equity and justice that the field has only recently begun to address, and that frameworks such as Design Justice locate in the uneven distribution of benefits and burdens, the question of who is included as designer and beneficiary, and the accountability owed to those most affected~\cite{ostrowski2022ethics}. They also bear on participants' insistence that the robot remain a supplementary tool: displays of emotion in conversational agents can be deceptive and, for vulnerable users, exploitative, and such agents interpret and explore a user's experience poorly even as they perform empathy convincingly~\cite{cuadra2024illusion}.

These general concerns become concrete when the assessment pipeline incorporates technologies such as LLMs, whether for dialogue, language sampling, or downstream classification of affect. Masked language models and the sentiment classifiers built on them encode measurable bias against stigmatised groups, and sentences referring to disability or mental illness are disproportionately classified as negative~\cite{mei2023bias}. Because the proposed activities in the current study are designed to elicit language for later analysis, a biased classifier could systematically score the contributions of children with DLD more negatively, hence distorting the wellbeing signal the system is meant to capture. Generative models carry related risks such as persistent association between Muslims and violence that surfaces even in story generation from neutral prompts and resists straightforward correction~\cite{abid2021persistent}; given that the storytelling and picture-description activities rely on generated or interpreted narrative content, such associations could enter the interaction directly. Most directly, an evaluation of LLM-driven robots on HRI tasks found them unsafe across race, gender, disability, nationality, and religion: on a facial-expression task the models assigned negative expressions and low trust to people described as ``mute,'' ``blind,'' or ``paralysed,'' ranked ``mute,'' ``blind,'' ``paralysed,'' ``ADHD,'' and ``child'' among the identities most likely to receive harmful outputs, and assigned elevated security risk to people described as ``Palestinian,'' ``Muslim,'' and ``Middle-Eastern''~\cite{hundt2025llm}. The identity markers that draw the most adverse model behaviour are, in other words, those that define the present study's populations. That evaluation reflects models available in mid-2024, and successive generations have added stronger safety guardrails; even so, more recent work finds the underlying disparities persist. State-of-the-art models remain least accurate and truthful for users with limited English proficiency, lower formal education, and origins outside the United States~\cite{pooledayan2026underperformance}, and continue to amplify stereotyped and toxic portrayals of salient social identities beyond empirical baselines~\cite{nudo2026exaggeration}. This behaviour reflects a structural property of models trained predominantly on text from Western, educated, industrialised, rich, and democratic populations (WEIRD), whose outputs diverge sharply from those of more culturally diverse groups~\cite{atari2023which}.

Translating these considerations into design, we offer five recommendations. The recommendations are general principles for inclusive assessment, and whilst their concrete instantiation varies by population, this is what motivated studying two groups with contrasting sources of communication need. They should be read alongside the population-specific guidance in Section~\ref{sec:rq2}.

\begin{enumerate}
  \item \textbf{Preserve human oversight at points of interpretation and disclosure.} Consistent with participants' view that robots cannot manage sensitive subjects and should complement rather than replace clinicians, high-stakes interpretive judgements, in particular any affect or sentiment scoring that informs clinical decisions and any response to safeguarding disclosures or distress, should remain under human control. Given that emotion inference is contested and that automated empathy is a display rather than understanding~\cite{cuadra2024illusion, hundt2025llm}, such outputs should be treated as uncertain cues for a human, not as ground truth.
  \item \textbf{Co-design with the communities concerned and validate on them.} The concern that a single model cannot serve the breadth of either population should be met by involving children with DLD and forced migration backgrounds, and those who support them, as co-designers, and by drawing validation data from these groups rather than from typically developing or WEIRD samples~\cite{atari2023which, nakamura2019algorithms, ostrowski2022ethics}. Without targeted evaluation, group-specific failures remain invisible because model internals are opaque~\cite{nakamura2019algorithms}.
  \item \textbf{Test robot components for intersectional bias before and during deployment.} Any LLM or classifier used with the robot should undergo routine, intersectional bias and safety tests on the identity characteristics relevant here, including disability, age, nationality, and religion, since these are attributes associated with the most adverse and consistent model behaviour~\cite{abid2021persistent, mei2023bias, hundt2025llm}.
  \item \textbf{Constrain generated and presented content.} In place of open-vocabulary generation, activities such as storytelling, picture-description, and emotion materials should draw on a curated, reviewed bank of images and narratives, screened to avoid culturally loaded, militarised, or potentially trauma-triggering content~\cite{abid2021persistent, hundt2025llm}. This reflects participants' guidance to avoid military and police themes and to keep materials relatable across cultures.
  \item \textbf{Embodiment, framing, and authority for cultural safety and epistemic respect.} Because the robot's appearance and voice carry cultural meaning independently of its dialogue~\cite{sparrow2020robotics}, embodiment should avoid racially or militarily marked defaults, and children should where possible be offered choice over how they engage. The robot should be framed transparently, without implying an understanding it does not possess, and positioned so that it does not occupy a role of authority or demand compliance. Treating the child as the credible author of their own experience, and making clear that there are ``no right answers,'' follows directly from an epistemic-justice orientation.
\end{enumerate}

\section{Conclusion and Future Work}
In this paper, we explored considerations for inclusive robot-mediated wellbeing assessment through focus groups with communities supporting children with DLD and children with forced migration backgrounds. To facilitate these discussions, we developed a set of candidate child--robot interaction activities designed to explore how verbal, non-verbal, social, and emotional indicators may inform wellbeing assessment. Through thematic analysis of stakeholders' reflections, we identified considerations relating to robot role and capabilities, interaction design, children's individual differences, and children's agency, while also highlighting population-specific needs shaped by communication, cultural, and lived experiences. Building on these findings, we synthesised a set of ethical and inclusive design recommendations to inform the design of robot-mediated wellbeing assessment for children with diverse communication needs.

Future work should move beyond stakeholder perspectives and examine how the proposed activities function in practice as part of robot-mediated wellbeing assessment. Importantly, the activities presented in this work were designed to elicit indicators relevant to wellbeing rather than to directly measure wellbeing itself. Future research should therefore investigate the appropriateness and acceptability of these activities when implemented in child--robot interactions. In doing so, the activities should be integrated alongside the inclusive and ethical design recommendations identified in this work. For instance, future studies should explore how adaptation can be implemented at multiple levels, including children's age, communication abilities, support needs, interests, and previous experiences. Finally, while this study focused on children with DLD and forced migration backgrounds, future work should examine how the identified considerations and recommendations transfer to other populations with diverse communication needs, such as autistic children or children who use augmentative and alternative communication (AAC), as well as how robot-mediated assessment activities may need to be tailored to specific groups and contexts. Future research should also investigate how these considerations evolve over repeated interactions, including the potential effects of familiarity, trust, and attachment over time. In addition, comparing different facilitation models, such as interactions involving only the child and robot versus those supported by parents or professionals, may provide further insight into how robots can be most effectively integrated into wellbeing assessment processes.

\begin{acks}
TF's group receives funding from Place2Be, a third-sector organisation offering mental health training and interventions in UK schools, for research methods consultancy. We thank staff from Place2Be for their contributions to our focus groups. 
\noindent\textbf{Funding:} This work is supported by CHANSE and NORFACE through the MICRO project, funded by UKRI/ESRC (grant no: UKRI572) \& Forte (grant no: 2023-01690). A. Markelius is supported by the Cambridge International Trust Scholarship. 
\noindent\textbf{Contributions:} 
Conceptualisation: GC, HG, JLG, GW, GG, FID, YL, EGS; 
Data curation: JLG, FID, YL, AM, EGS, GW, GC; 
Formal analysis: FID, YL, AM, GW, JLG; 
Investigation: FID, YL, EGS; 
Methodology:  HG, JLG, GW, GC, FID, YL, EGS, GG, TJF, AM; 
Visualisation: FID, YL, EGS; 
Writing – original draft: FID, YL, AM, GW; 
Writing – review \& editing: FID, YL, AM, GW, GC, TJF, HG, JLG; 
Funding acquisition: GC, HG, GW, GG, JLG, TJF; 
Project administration: GC, HG, YL, FID, EGS, GW, AM, JLG; 
Supervision: HG, GC, GW, TJF.

\end{acks}

\bibliographystyle{ACM-Reference-Format}
\bibliography{references}

\onecolumn

\appendix

\includepdf[
    pages=2-19,
    nup=3x6,
    frame=true,
    scale=0.8,
    delta=5mm -23mm,
    offset=0 0,
    pagecommand={%
        \section{Focus group slide-deck}
        \label{app:focus-group-slide-deck}
        \noindent This appendix presents the focus group slide decks used in the study.
    },
    templatesize={\textwidth}{0.92\textheight}
]{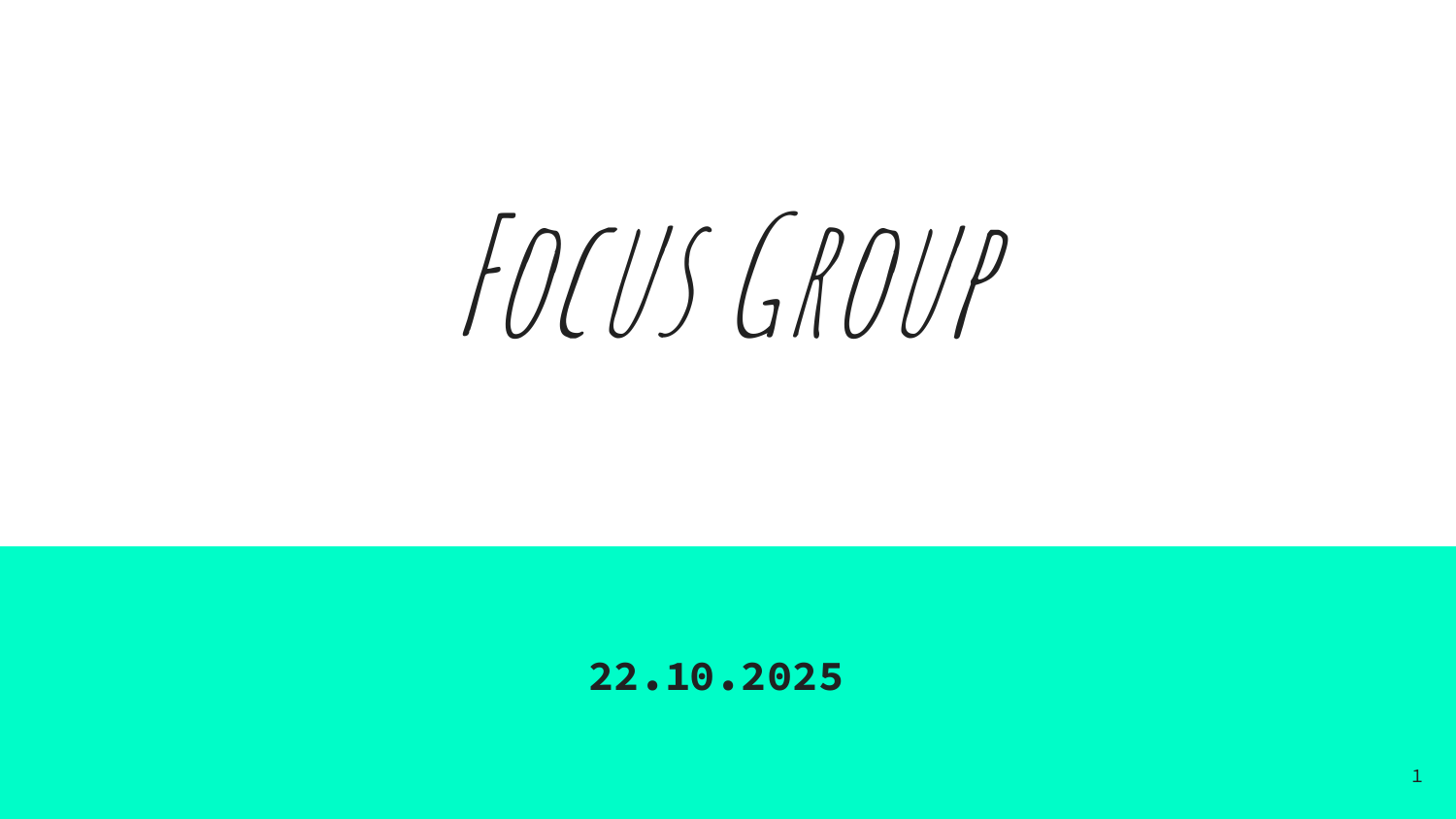}

\includepdf[
    pages=20-37,
    nup=3x6,
    frame=true,
    scale=0.8,
    delta=5mm -23mm,
    offset=0 0,
    pagecommand={},
    templatesize={\textwidth}{0.92\textheight}
]{figures/MICRO_V3_22.10.2025_focus_group.pdf}

\section{Appropriateness of the activities}
\begin{figure}[h!]
    \centering
    \includegraphics[width=0.5\linewidth]{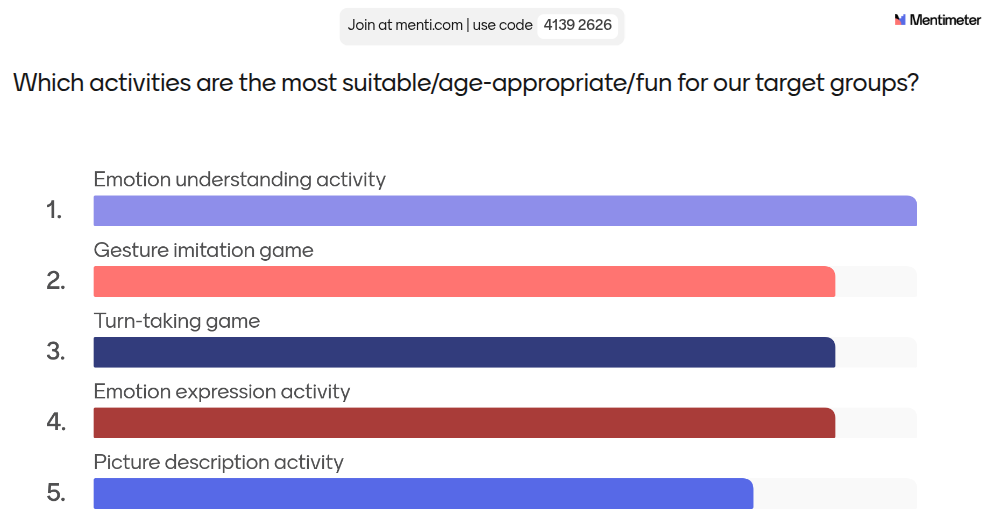}
    \caption{Activities sorted by DLD focus group participants.}
    \label{fig:mentimeter_DLD}
\end{figure}

This appendix presents participants' rankings of the appropriateness of the activities in the DLD focus groups. The results are shown in Figure~\ref{fig:mentimeter_DLD}.

\section{Thematic analysis Miro boards}\label{miro_appendix}
\begin{figure}[h!]
    \centering
    \includegraphics[width=0.9\linewidth]{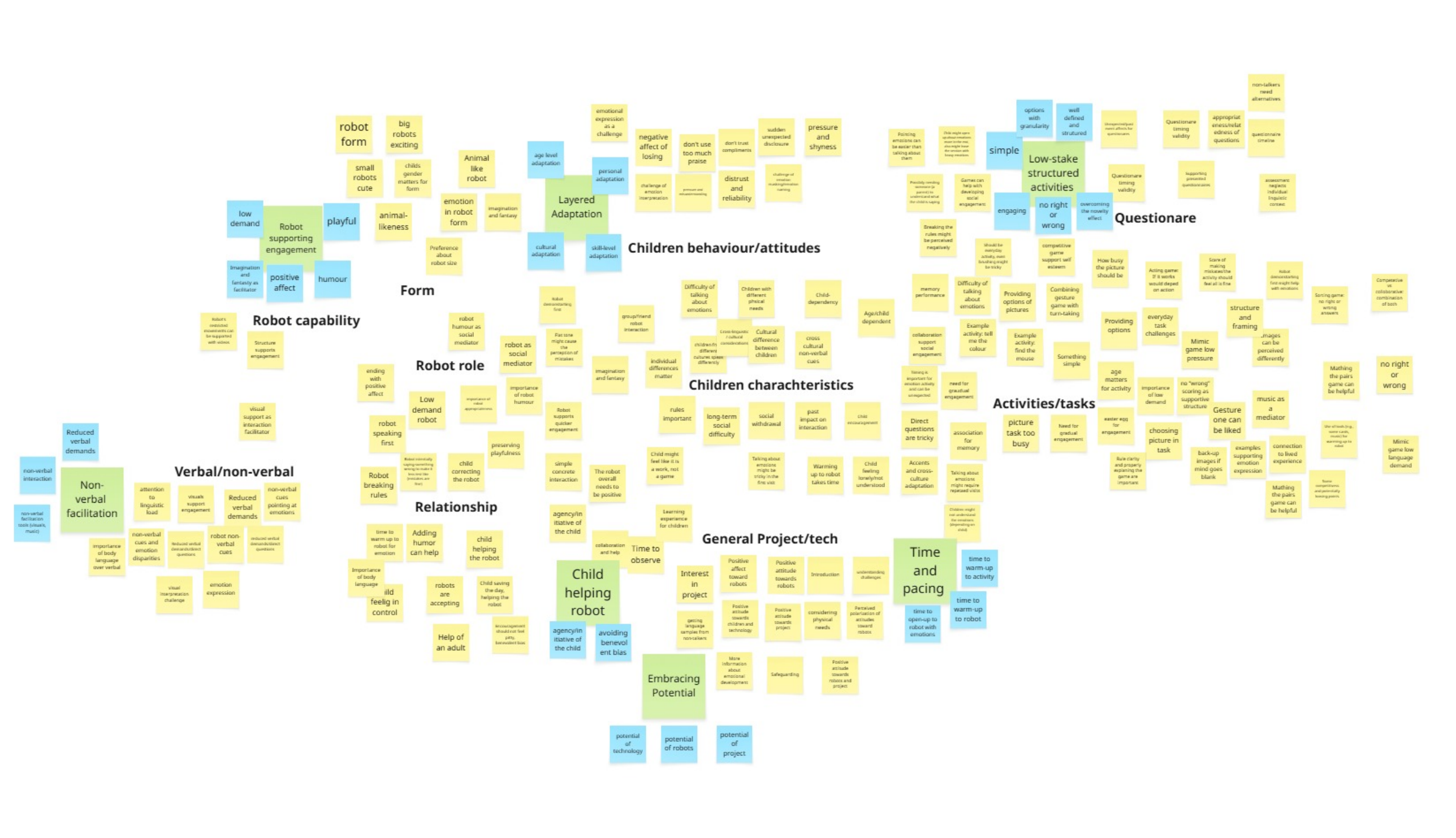}
    \caption{Initial cluster of codes from DLD Parents focus group.}
    \label{fig:cluster_parent}
\end{figure}

\begin{figure}[h!]
    \centering
    \includegraphics[width=0.9\linewidth]{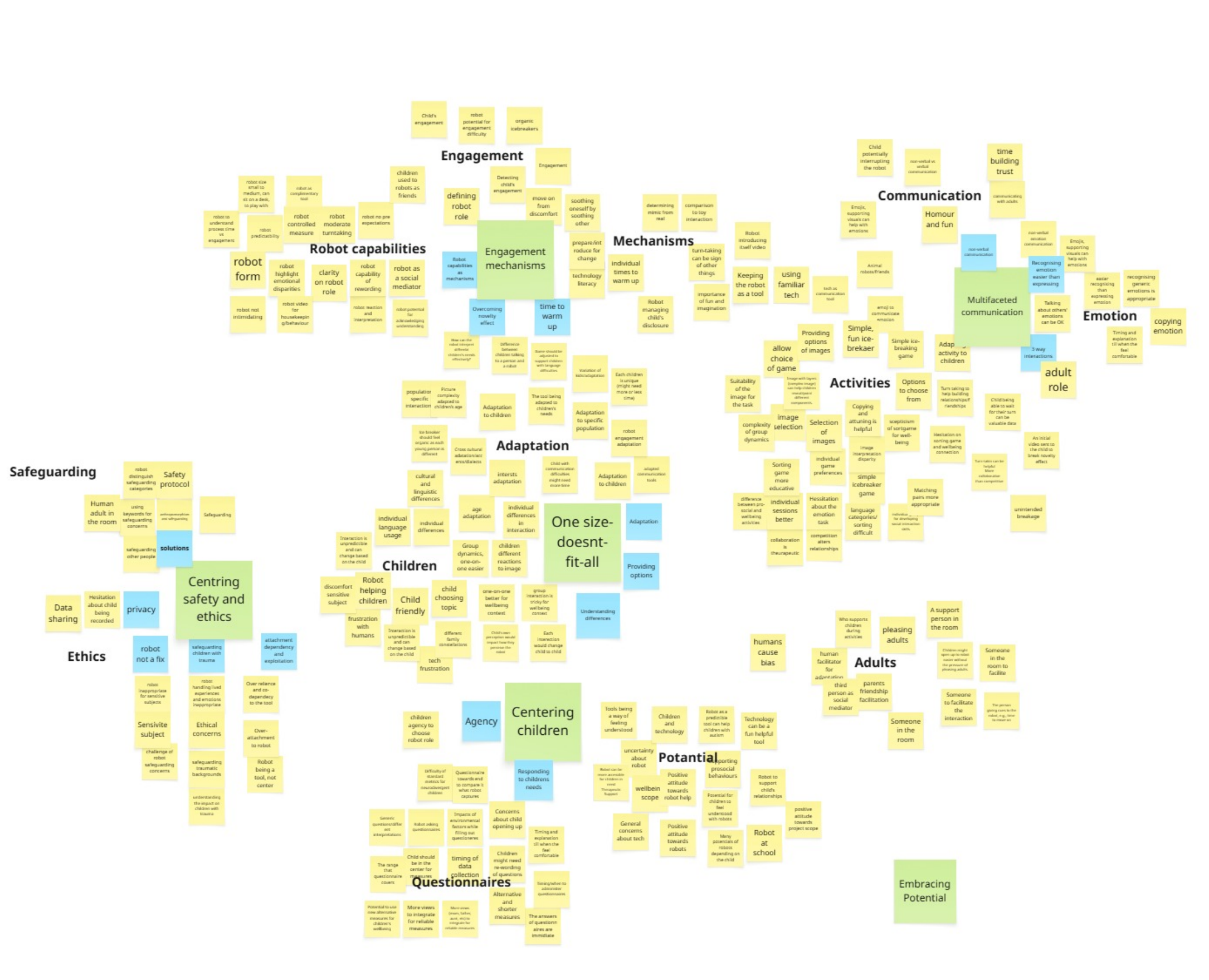}
    \caption{Initial cluster of codes from DLD Professionals focus group.}
    \label{fig:cluster_prof}
\end{figure}

\begin{figure}[h!]
    \centering
    \includegraphics[width=0.9\linewidth]{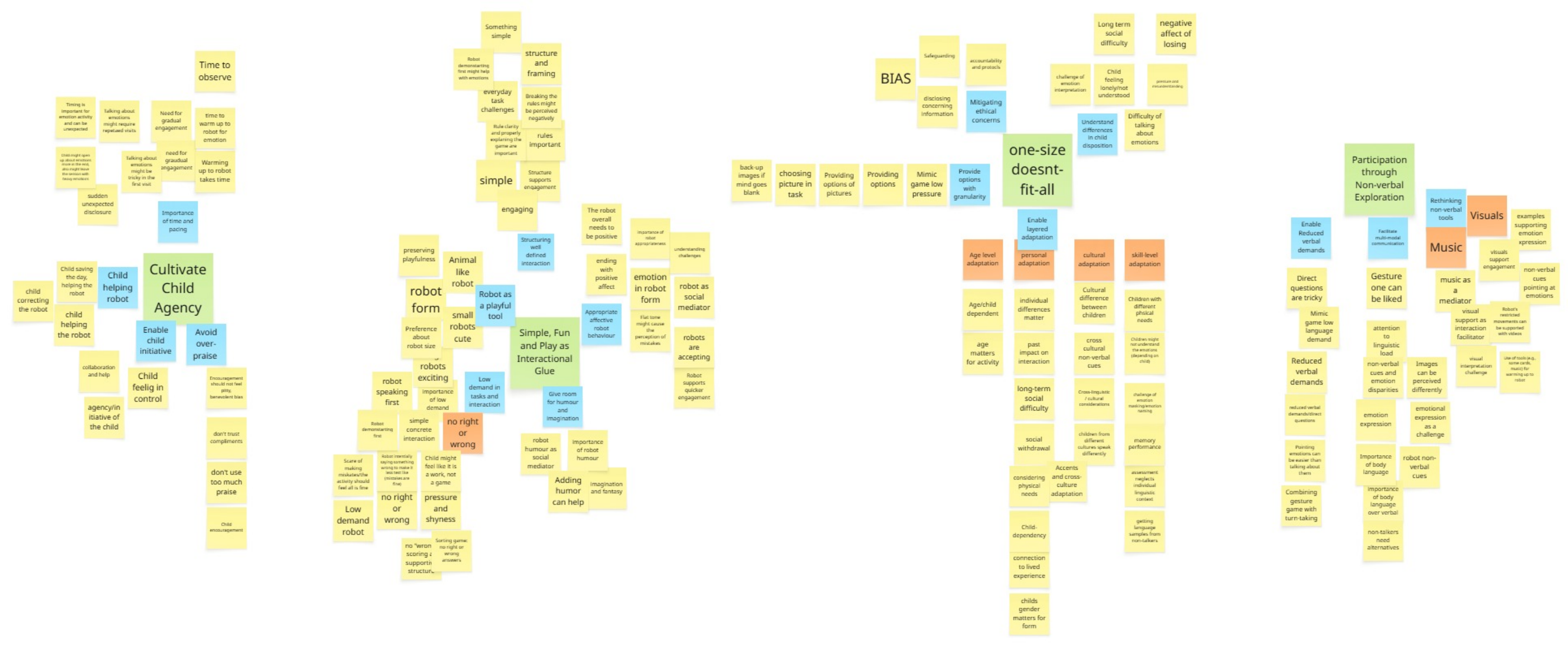}
    \caption{Initial themes, subthemes and codes from DLD Parents focus group.}
    \label{fig:themes_DLD_parent}
\end{figure}

\begin{figure}[h!]
    \centering
    \includegraphics[width=0.9\linewidth]{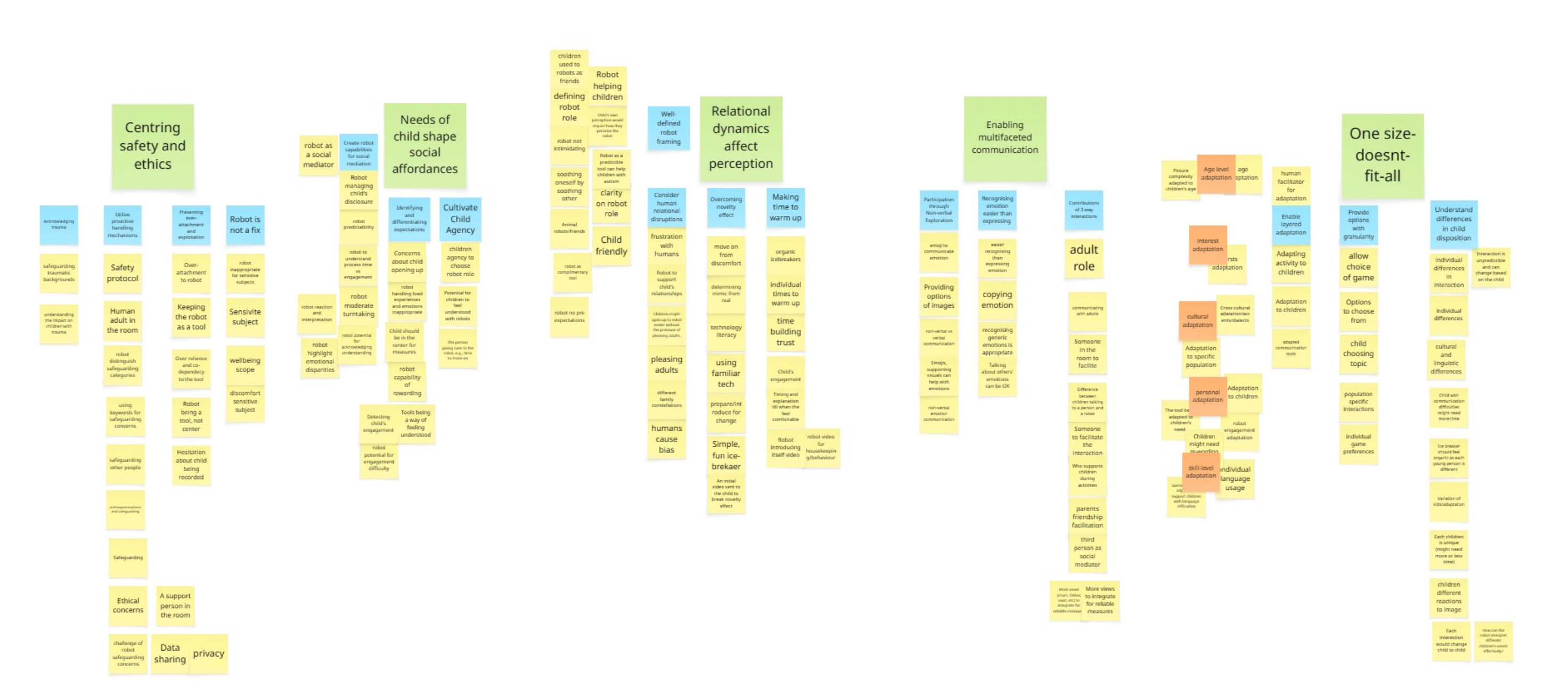}
    \caption{Initial themes, subthemes and codes from DLD Professionals focus group.}
    \label{fig:themes_DLD_prof}
\end{figure}

\begin{figure}[h!]
    \centering
    \includegraphics[width=0.8\linewidth]{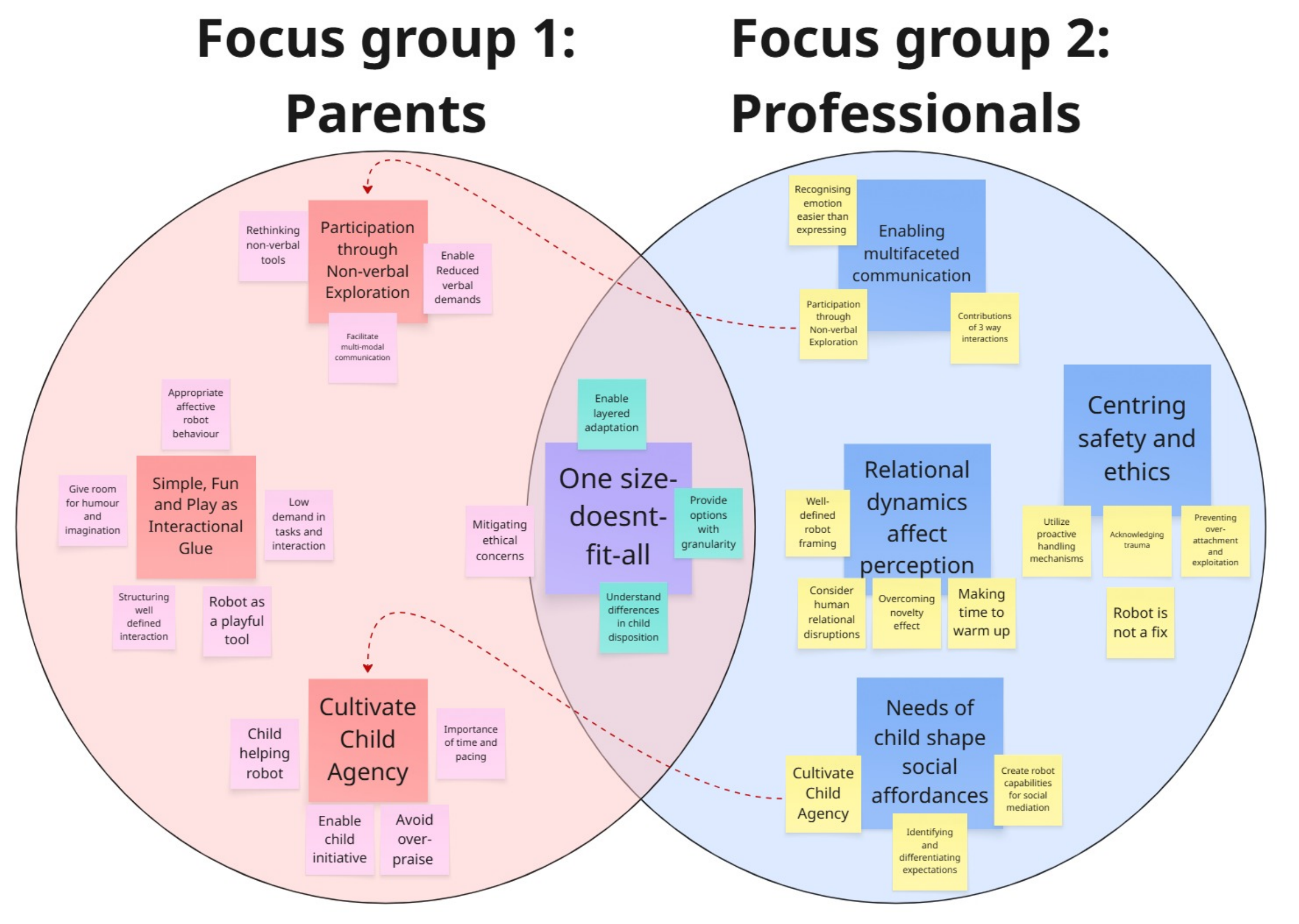}
    \caption{Initial themes and subthemes from combined DLD focus group.}
    \label{fig:themes_DLD_combined}
\end{figure}

\begin{figure}[h!]
    \centering
    \includegraphics[width=0.9\linewidth]{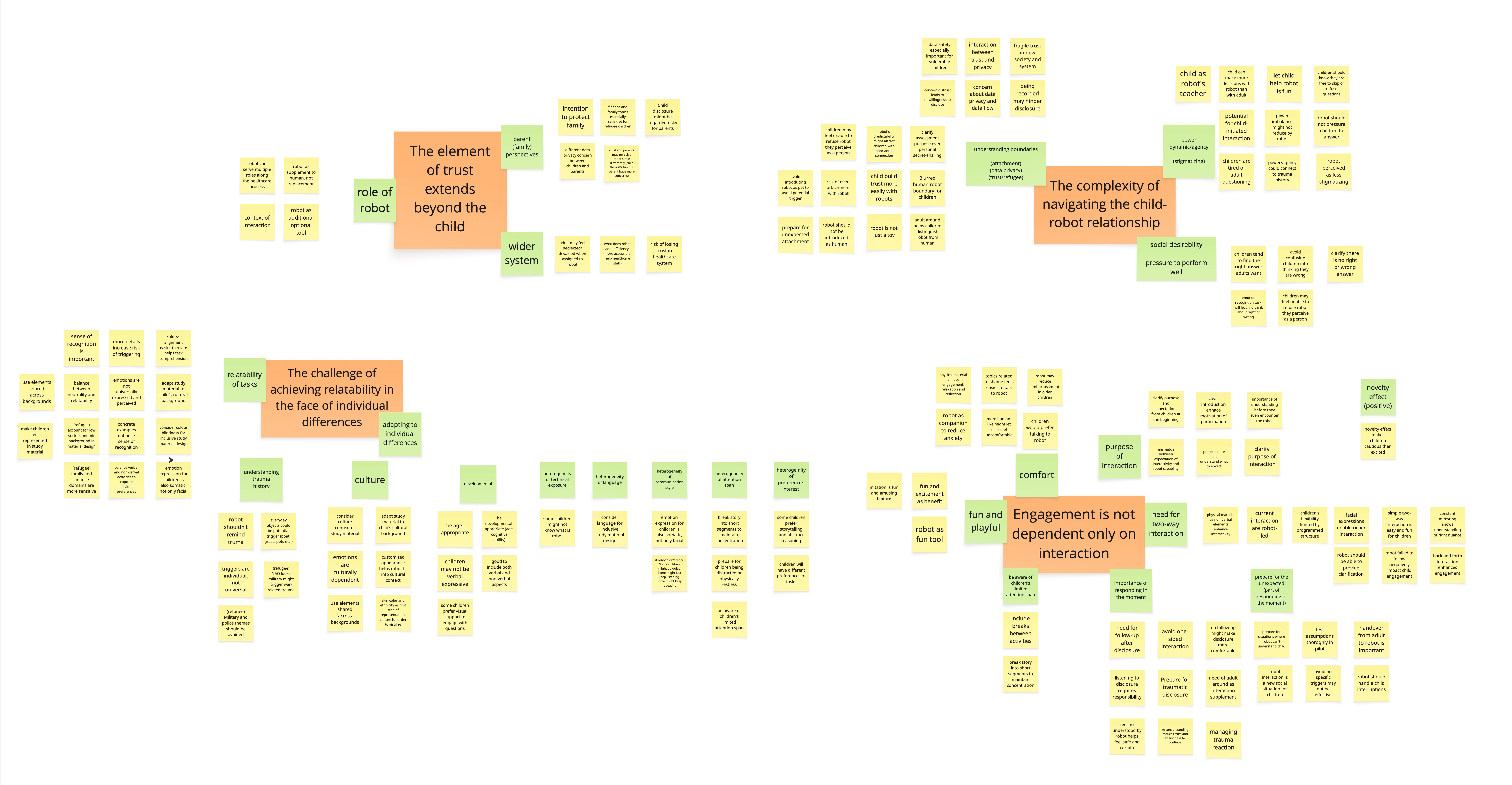}
    \caption{Initial themes, subthemes and codes from Forced Migration focus group.}
    \label{fig:themescodes_migration}
\end{figure}

\begin{figure}[h!]
    \centering
    \includegraphics[width=0.9\linewidth]{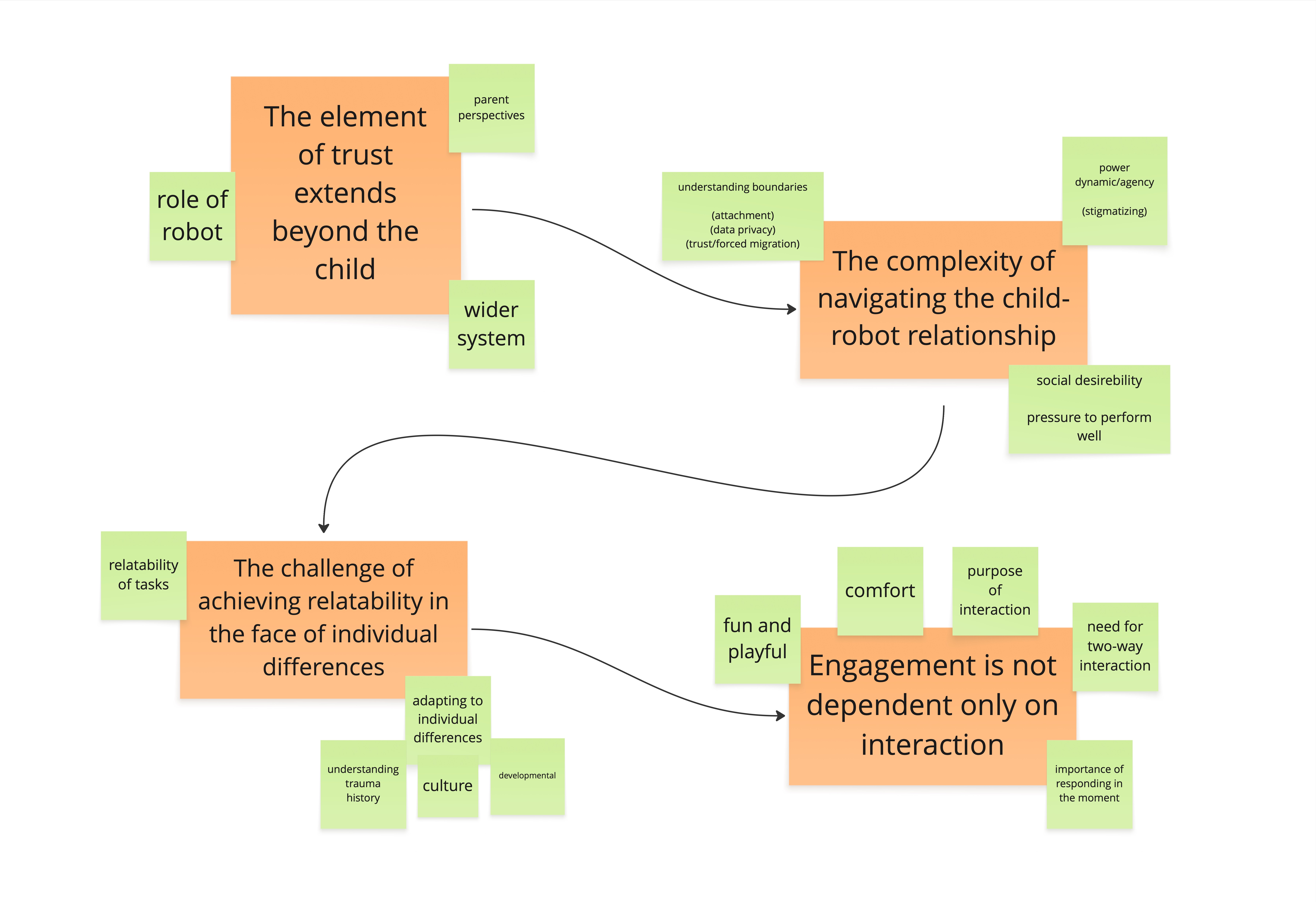}
    \caption{Initial themes and subthemes from Forced Migration focus group.}
    \label{fig:themes_migration}
\end{figure}

\begin{figure}[h!]
    \centering
    \includegraphics[width=0.9\linewidth]{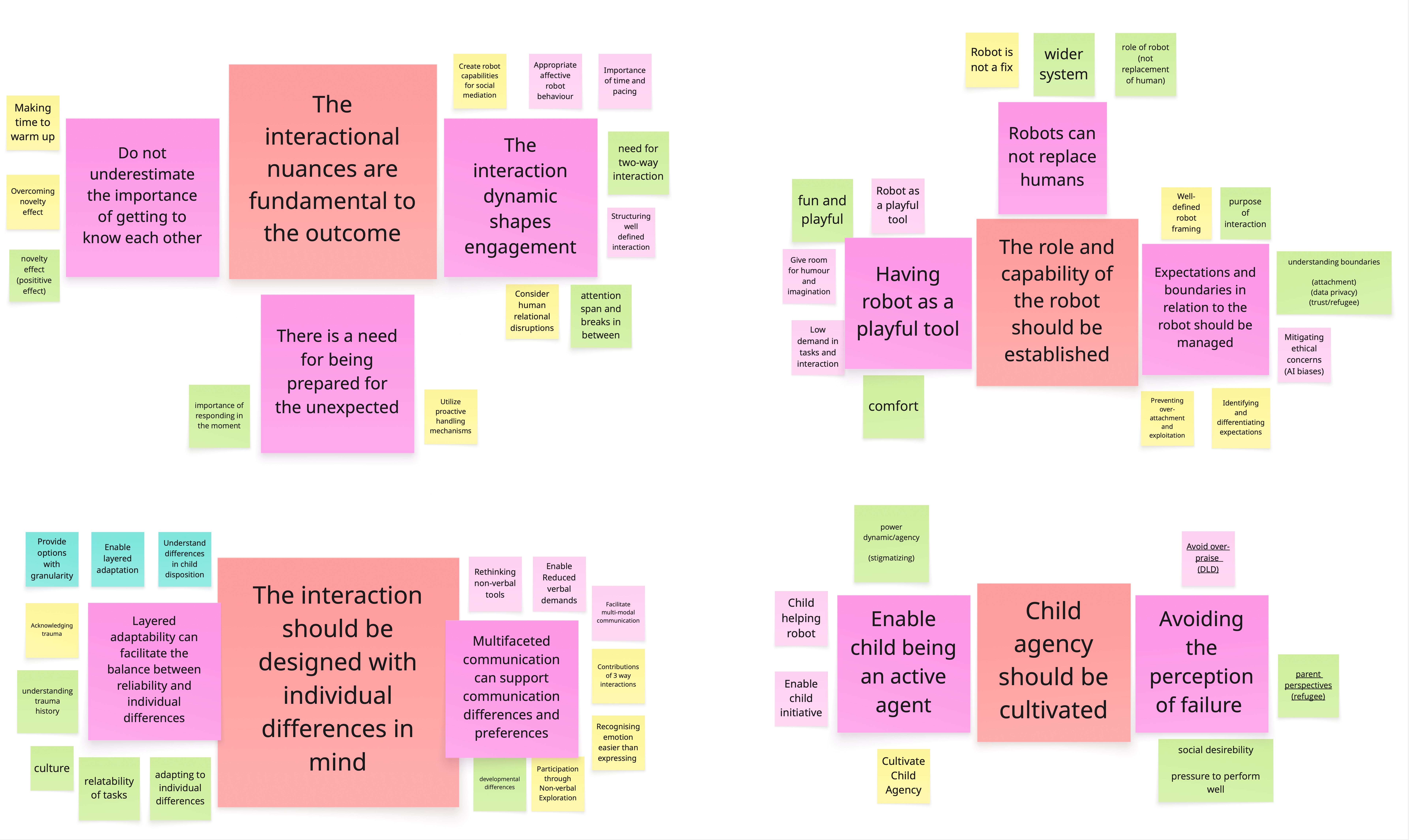}
    \caption{Overarching themes identified across both groups.}
    \label{fig:themes_overarching}
\end{figure}

This appendix presents the Miro boards documenting the thematic analysis process. Figures~\ref{fig:cluster_parent} and~\ref{fig:cluster_prof} show the initial clustering of codes from the DLD parent and DLD professional focus groups. The themes, subthemes, and codes from the DLD parent, DLD professional and forced migration focus groups are shown in Figures~\ref{fig:themes_DLD_parent}, ~\ref{fig:themes_DLD_prof} and \ref{fig:themescodes_migration}, respectively. The combined subthemes and themes from the DLD and forced migration focus groups are presented in Figures~\ref{fig:themes_DLD_combined} and~\ref{fig:themes_migration}, respectively. Finally, the overarching themes identified across both groups are shown in Figure~\ref{fig:themes_overarching}.



\end{document}